\documentclass[10pt,twocolumn,letterpaper]{article}

\usepackage{iccv}
\usepackage{times}
\usepackage{epsfig}
\usepackage{graphicx}
\usepackage{amsmath}
\usepackage{amssymb}

\usepackage{siunitx}        %
\usepackage{booktabs}
\usepackage{multirow}
\usepackage{colortbl}		%
\usepackage{placeins}		%
\usepackage[table,xcdraw]{xcolor}
\usepackage[pagebackref=true,breaklinks=true,letterpaper=true,colorlinks,bookmarks=false]{hyperref}

\DeclareMathOperator{\KL}{KL}

\newcommand{\pdata}{p_\mathcal{X}}
\newcommand{\ptriplet}{p_\mathbb{T}}
\newcommand{\pose}{\pi}
\newcommand{\app}{\alpha}

\newcommand{\idxapp}{1}
\newcommand{\idxout}{2}
\newcommand{\idxpose}{3}
\newcommand{\xpose}{\ensuremath{x_{\idxpose}}\xspace}
\newcommand{\xapp}{\ensuremath{x_{\idxapp}}\xspace}
\newcommand{\xout}{\ensuremath{x_{\idxout}}\xspace}

\newcommand{\Lall}{\mathcal{L}}
\newcommand{\Lrec}{\mathcal{L}_{\text{rec}}}
\newcommand{\Lvb}{\mathcal{L}_{\text{VB}}}
\newcommand{\Lmi}{\mathcal{L}_{\text{MI}}}

\DeclareSIUnit{\px}{px}

\newcommand{\citep}{\cite}
\newcommand\blfootnote[1]{%
  \begingroup
  \renewcommand\thefootnote{}\footnote{#1}%
  \addtocounter{footnote}{-1}%
  \endgroup
}

\iccvfinalcopy %

\def\iccvPaperID{3555} %

\ificcvfinal\fi
\begin{document}

\title{Unsupervised Robust Disentangling of \\Latent Characteristics for Image Synthesis}

\author{Patrick Esser, \space Johannes Haux, \space Bj\"orn Ommer\\
  Heidelberg Collaboratory for Image Processing\\
  IWR, Heidelberg University, Germany\\
  {\tt\small firstname.lastname@iwr.uni-heidelberg.de}
}

\maketitle
\thispagestyle{empty}

\begin{abstract}
Deep generative models come with the promise to learn an explainable
representation for visual objects that allows image sampling, synthesis,
and selective modification. The main challenge is to learn to properly model the
independent latent characteristics of an object, especially its appearance
  and pose. We present a novel approach that learns disentangled
  representations of these characteristics and explains them individually.
Training requires only pairs of images depicting the same object appearance,
but no pose annotations.
We propose an additional classifier that estimates the minimal amount of
  regularization required to enforce disentanglement. Thus both
  representations together can completely explain an image while being
  independent of each other. Previous methods based on adversarial
  approaches fail to enforce this independence, while methods based on
  variational approaches lead to uninformative representations.
In experiments on diverse object categories, the approach
successfully recombines pose and appearance to reconstruct and retarget
novel synthesized images. We achieve significant improvements over
state-of-the-art methods which utilize the same level of supervision, and
reach performances comparable to those of pose-supervised approaches.
  However, we can handle the vast body of articulated object classes for
  which no pose models/annotations are available.
\end{abstract}

\section{Introduction}

Supervised end-to-end training on large volumes of tediously labelled data
has tremendously propelled deep learning \cite{alexnet}. The discriminative
learning paradigm has enabled to train deep network architectures with
millions of parameters to address important computer vision tasks such as
image categorization \cite{simonyan2014very}, object detection \cite{ren2015faster}, and segmentation
\cite{ronneberger2015u}. The network architectures underlying these discriminative models
have become increasingly complex \cite{vgg} and have tremendously increased
in depth \cite{resnet} to yield great improvements in performance. However,
the ability to explain these models and their decisions suffers due to this
discriminative end-to-end training setup \cite{zeiler_fergus,advsampl,distill}.

\begin{figure}[h]
        \centering
        \includegraphics[width=0.9\linewidth]{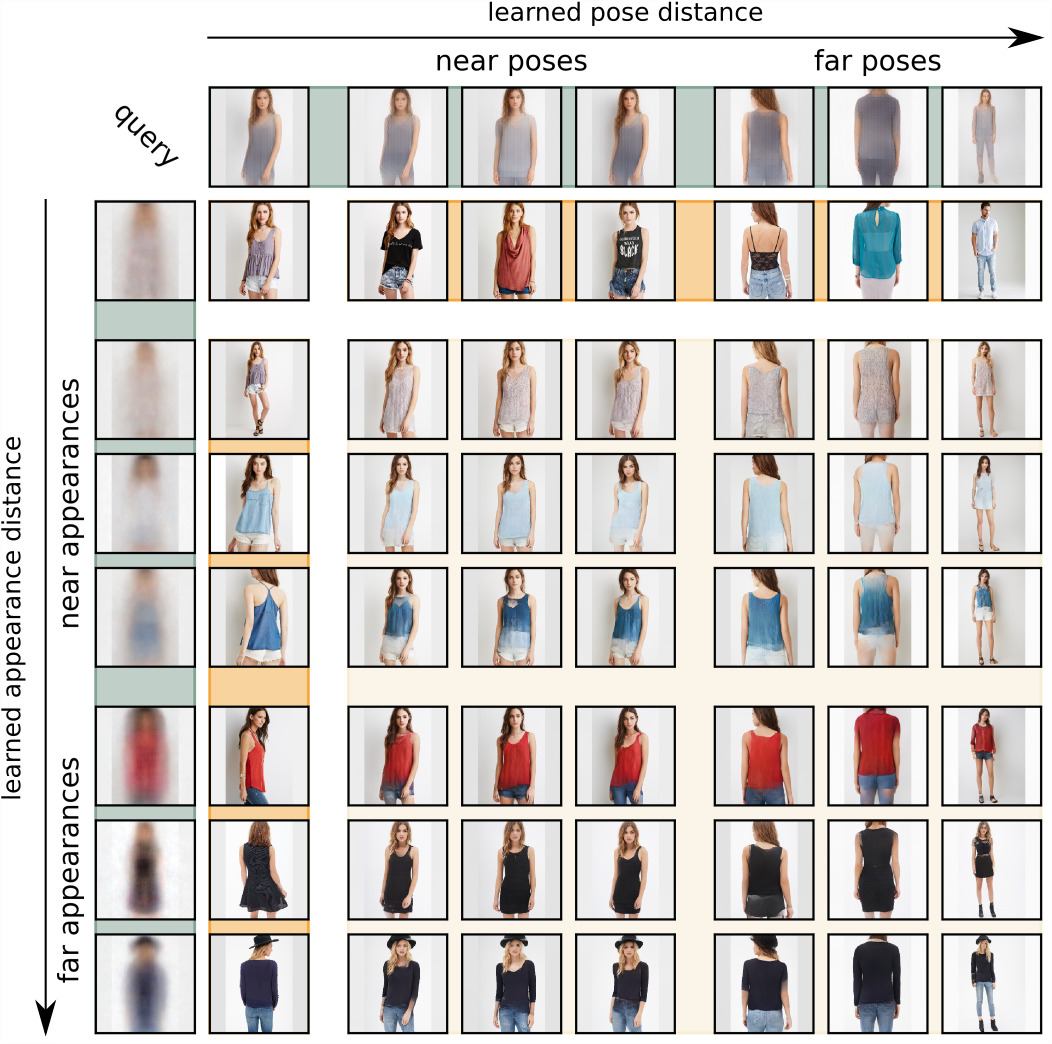}
        \caption{Without annotations about pose, we disentangle images into
        two independent factors: pose and appearance. Starting from a query
        image (top left), we can extract and visualize its pose
        representation (top row), and retrieve images based on their pose
        similarity. The visualizations (Sec.~\ref{visualizations}) in the
        first row show that pose is learned accurately and contains no
        information about appearance.  Similarly, appearance is visualized
        in the first column. Because our representations are both
        independent and informative, we can recombine arbitrary combinations
        to synthesize what an appearance would look like in a specific pose.
        More results can be found at
        \href{https://compvis.github.io/robust-disentangling}{https://compvis.github.io/robust-disentangling}.}
        \label{fig:rmatrix}
\end{figure}

Consequently, there has recently been a rapidly increasing interest in deep
generative models \cite{VAE,VAE2,gan,ar}. These aim for a complete
description of data in terms of a joint distribution and can, in a natural
way, synthesize images from a learned representation. Thus, besides
explaining the joint distribution of all data, they provide a powerful tool
to visualize and explain complex models \cite{chen2016infogan}. 

However, while already simple probabilistic models may produce convincing
image samples, their ability to describe the data is lacking. For
instance, great progress in image synthesis and interpolation between
different instances of an object category (e.g., young versus old faces) has
been achieved \cite{Weinberger,lample2017fader,attgan}. But these models
explain all the differences between instances as a change of appearance.
Consequently, changes in posture, viewpoint, articulation and the like
(subsequently simply denoted as pose) are blended with changes in color or
texture.

\begin{figure}
        \centering
        \includegraphics[width=\linewidth]{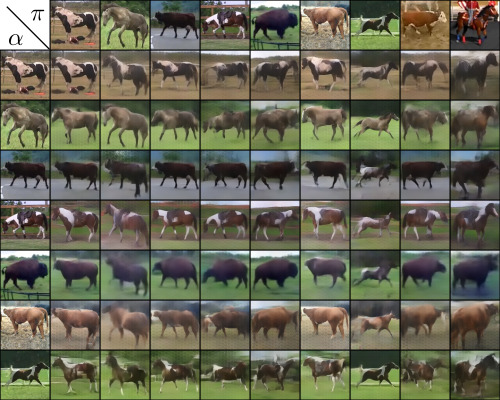}
        \caption{First row: pose target. First column: appearance target.
        Because our method does not require keypoint-annotations or class
        information, it 
        can be readily applied on video datasets
        \cite{got10k,lasot}. Besides intra-species analogies, our approach can
        also imagine inter-species analogies: How does a cow look like in a
        pose specified by a horse?}
        \label{fig:chmatrix}
\end{figure}

To address the different characteristics of pose and appearance, many
recent approaches started to rely on existing discriminative pose detectors.
While these models show good results on disentangled image generation, their
applicability is limited to domains with existing, robust pose detectors.
This introduces two problems: The output of the pose detector introduces
a bias into the notion of what constitutes pose, and labeling large scale
datasets for each new category of objects to be explained is unfeasible.
How can we learn pose and appearance without these problems?
The task naturally calls for two encoders \cite{denton2017unsupervised,hadad,mathieu}
to extract representations of
appearance and pose, and a decoder to reconstruct an image from them. To
train the model, one encoder infers the pose from the image to be
reconstructed; the other encoder infers the appearance from another image
showing the same appearance. Without further constraints, the reconstruction
task alone produces a degenerate solution \cite{cyclevae,szabo,mathieu}: The pose encoding
contains all the information---including appearance---and the decoder
ignores the appearance representation. In this case, the model collapses to
an autoencoder and avoiding this is the main goal of disentanglement.

There are two principle approaches to disentanglement. Variational approaches
\cite{kingmasemi,cyclevae} utilize a stochastic representations that is regularized towards a
prior distribution with the Kullback-Leibler (KL) divergence.
This regularization penalizes information in the representation and---for
large regularization weights---encourages disentanglement \cite{betavae}.
However, both entangled and disentangled content in the representation is
penalized by the same amount. Disentanglement therefore comes at the price of
uninformative representations: Reconstructions are blurry and the
representations cannot explain the complete image.

\begin{figure}
        \centering
        \includegraphics[width=\linewidth]{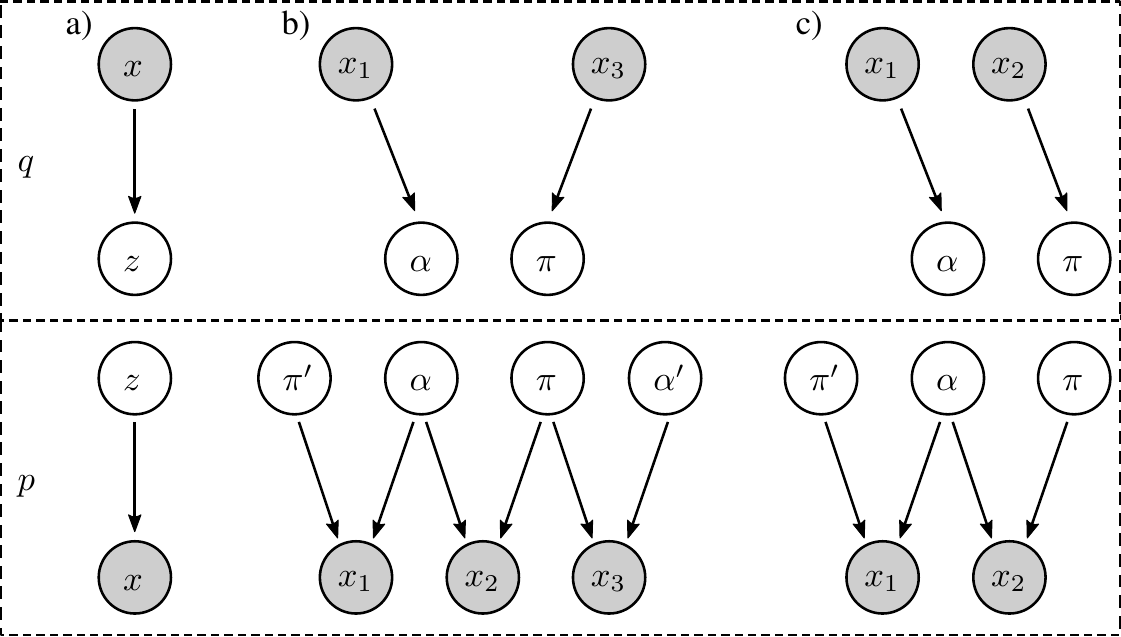}
        \caption{Graphical models describing the dependencies within a) a simple
    latent variable model b) a complete model of disentangled factors c) the
    same model with unobserved $x_3$. $p$ describes the generative
    process; $q$ describes a variational approximation of the inference process.}
        \label{fig:probmodel}
\end{figure}

\newcommand{\daleft}[1]{\includegraphics[width=0.0516\linewidth]{assets/comparematrix/#1}}
\newcommand{\dafig}[1]{\includegraphics[width=0.155\linewidth]{assets/comparematrix/#1}}
\newcommand{\dahead}[1]{\includegraphics[width=0.155\linewidth]{assets/comparematrix/#1}}
\begin{table*}[ht]
\begin{tabular}{lccccc}
  & a) only variational & b) only adversarial & c) equal combination & d) fixed
  combination & e) adaptive combination \\
 & \dafig{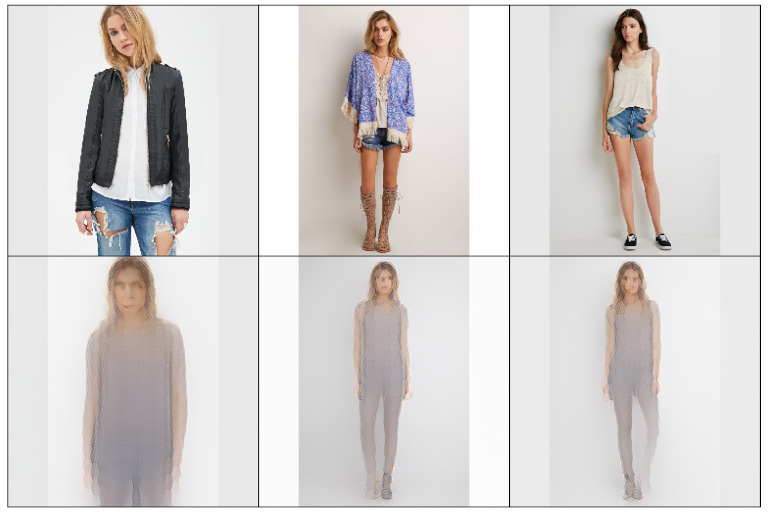} & \dafig{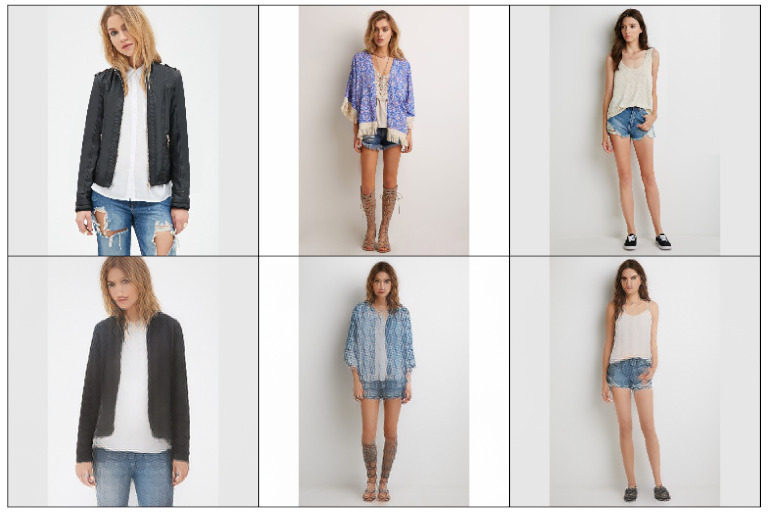} &
\dafig{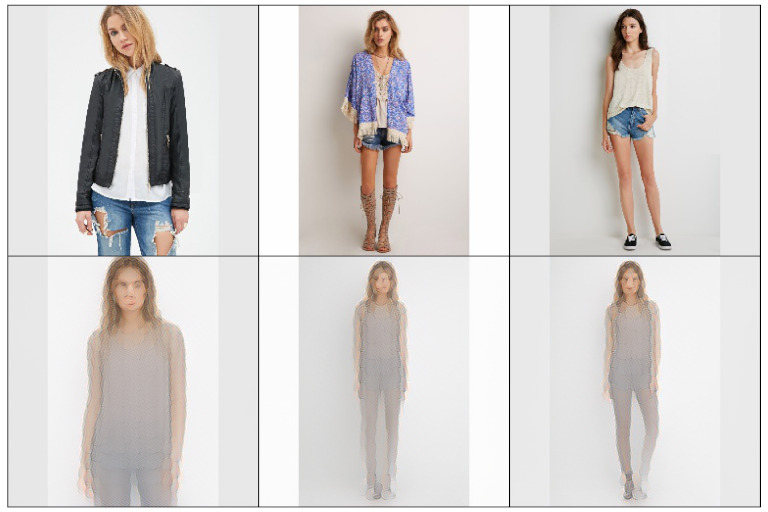} &
  \dafig{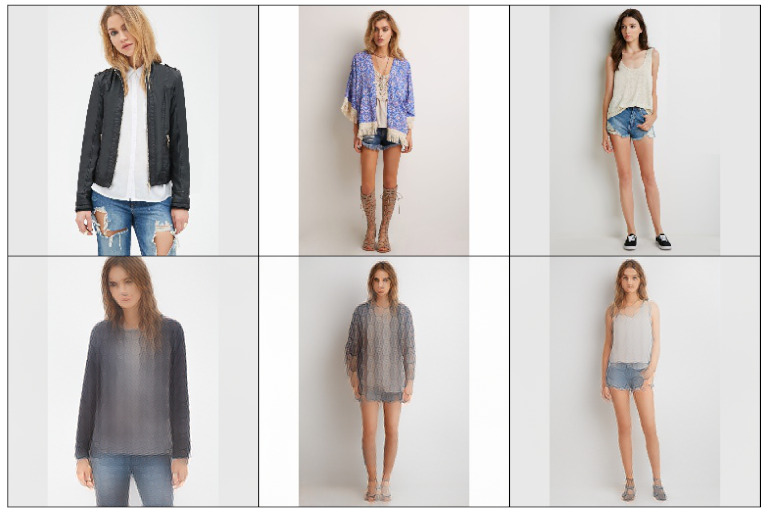} & \dafig{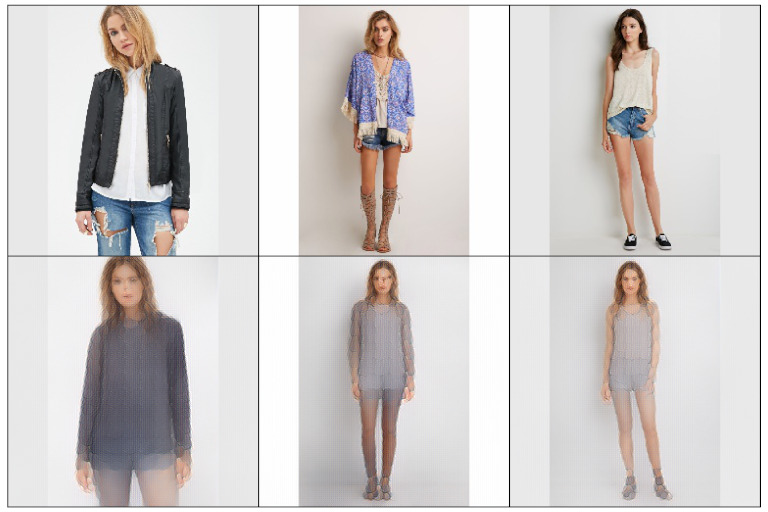} \\
\daleft{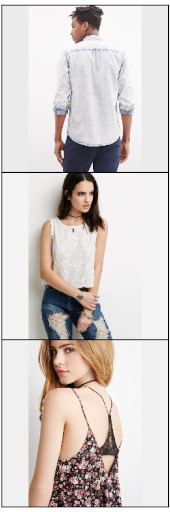}
 & \dafig{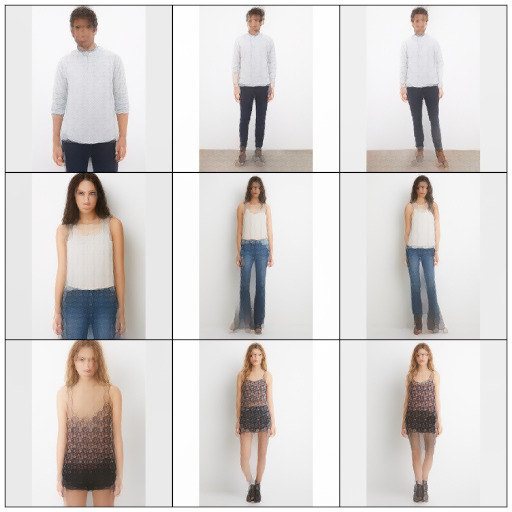} & \dafig{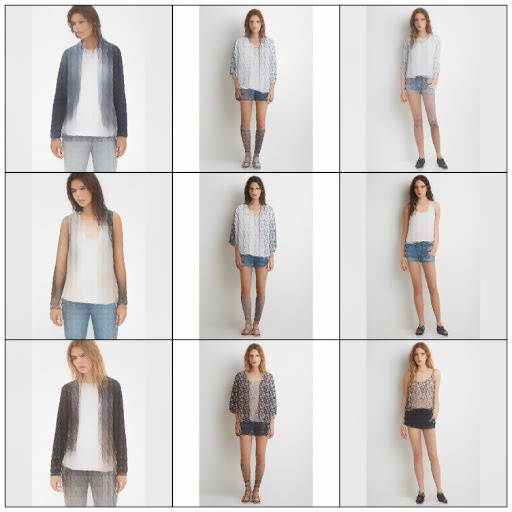} &
\dafig{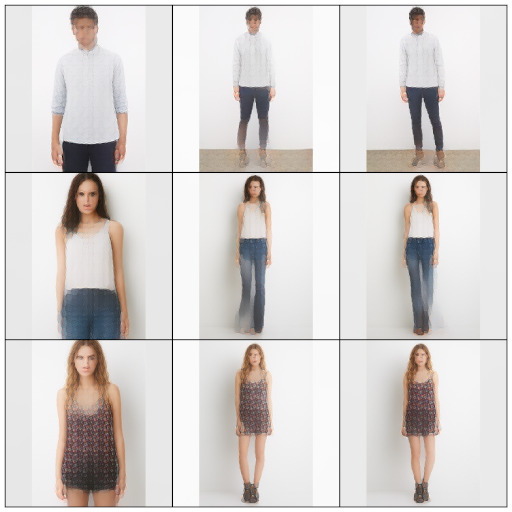} &
  \dafig{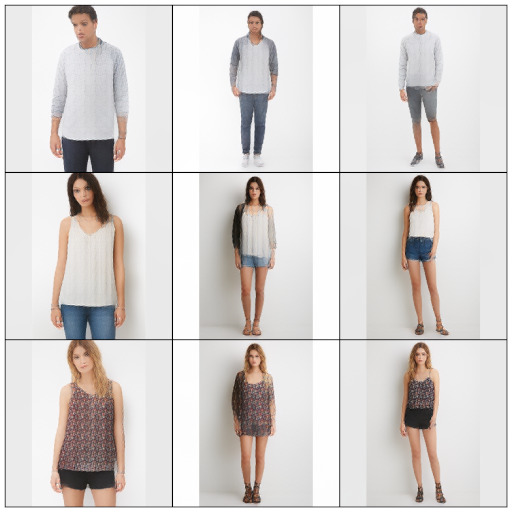} & \dafig{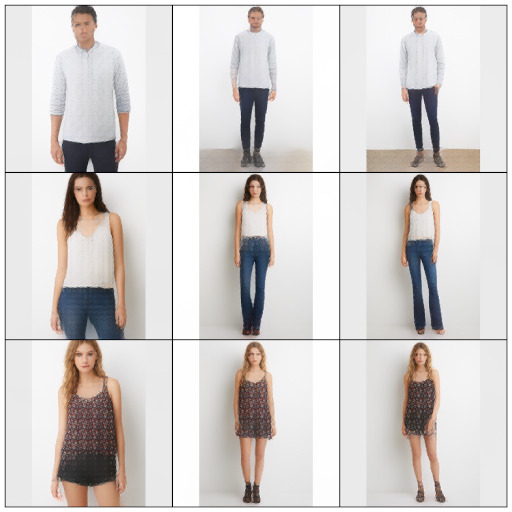} \\
  \midrule
  $I_T$ & 0.1255 & 0.1254 & 0.0919 & 0.1258 & 0.1225 \\
  \midrule
$I_{T'}$ & \cellcolor[HTML]{dfffdf}0.1239 & \cellcolor[HTML]{ffdddb}0.8727 & \cellcolor[HTML]{dfffdf}0.1247 & \cellcolor[HTML]{ffdddb}0.4096 & \cellcolor[HTML]{dfffdf}0.1201 \\
  \midrule
$\Lrec$ & 3.9350 & 2.9818 & 3.9615 & 3.2504 & \textbf{3.5279} \\
  \midrule
\end{tabular}
\caption{First row: Pose targets. Second row: Pose visualizations. First
  column: Appearance targets. Enforcing a MI constraint of $\epsilon = 0.125$ in
  eq.~\eqref{objective2}:
  a) leads to lossy representations. Pose is not accurately captured,
  leading to blurry synthesis results and high
  reconstruction loss.
  b) $T$ indicates successful disentanglement but
  $T'$ reveals high entanglement. Visualizations show that pose contains
  complete appearance information and the transfer task fails.
  c) Same problems as a) because disentanglement relies again on the variational approach.
  d) Improved compared to b) but still fails at the transfer task.
  e) Our adaptive combination achieves
  disentanglement and accurate pose representations. We obtain the lowest
  reconstruction error of all the methods which can enforce the
  disentanglement constraint (shaded green). See also Sec.~\ref{visualizations}.}
\label{tab:abl}
\end{table*}

Adversarial approaches \cite{denton2017unsupervised,hadad,lample2017fader} to disentanglement have the potential to
provide a more fine-grained regularization. In these approaches a
discriminator estimates entanglement and its gradients are used to guide the
representations directly towards disentanglement. Therefore, they come with
the promise to selectively penalize nothing but entangled content. However,
we identify as the \textbf{key problem} that the encoder, having access to
the discriminator's gradients, learns to produce entangled representations
which are classified as disentangled. In contrast to adversarial attacks
on image classifiers \cite{aattack},
in our case, the attack happens at the level of representations instead of
images and implies that one cannot rely on adversarial approaches directly.

\textbf{Our contribution} is an approach for making adversarial approaches robust to
overpowering without being affected by uninformative representations.
We use a second classifier---whose gradients are never provided to the
encoder---to detect overpowering: A large difference in both entanglement
estimates implies that the first classifier is being tricked by the encoder.
However, to achieve disentangled representations, we cannot directly utilize
feedback from the second classifier, because this would reveal its gradients
to the encoder and make it vulnerable to being overpowered, too. Instead, we
use it indirectly to estimate the weight of a KL regularization term---we
increase it when overpowering is detected and decrease it otherwise. This
way, disentanglement comes from the first classifer, which is controlled by
the second.

\section{Disentanglement in probabilistic models}

\subsection{Latent variable models}
Let $x$ denote an image from an unknown data distribution $\pdata(x)$.
Probabilistic approaches to image synthesis approximate the unknown
distribution using a model distribution $p(x)$. To fit this distribution to
the data distribution, the maximum log-likelihood of data samples is
maximized:
\begin{equation}
  \label{ML}
  \max_p \mathbb{E}_{x \sim \pdata(x)} \log p(x)
\end{equation}

The distribution $p$ can then be sampled to
synthesize new images. To model $p$, latent variable
models assume that images are generated due to an underlying latent variable $z$
which is not observed. The full model distribution
\begin{equation}
  p(x, z) = p(x \vert z) p(z)
\end{equation}
is then specified in terms of the factors $p(x \vert z)$, which are typically
parameterized by a function class such as neural networks, and the factor
$p(z)$ which specifies a prior on the latent variable, typically given by
a simple distribution like a normal distribution.

A Generative Adversarial Network (GAN) \cite{gan} learns such a
model using density ratio estimation. The training algorithm can be
described as a two player game: A classifier tries to distinguish
between generated and real images; a generator tries to generate images
that are indistinguishable from real images. Because no inference process is
learned, they cannot explain existing images.

\begin{figure*}
    \centering
    \def\svgwidth{1\linewidth}
    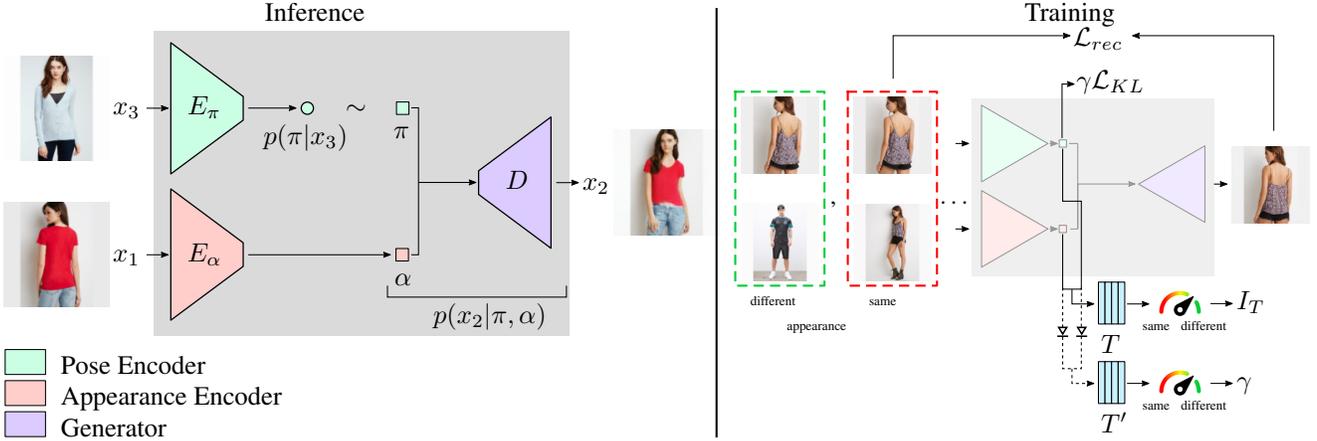
    \caption{Left: synthesizing a new image $\xout$ that exhibits the
    apperance of $\xapp$ and pose of $\xpose$. Right: training our
    generative model using pairs of images with same/different appearance.
    $T$ and $T'$ estimate the mutual information between
    $\pose$ and $\app$. The gradients of $T$ are used to guide
    disentanglement, while $T'$ detects overpowering of $T$ and estimates
    $\gamma$ to counteract it.}
    \label{fig:model}
\end{figure*}

The Variational
Autoencoder (VAE) \cite{VAE,VAE2} learns a latent variable model
using variational inference and the reparameterization trick.
The structure of the joint $p(x,z)$ and the corresponding encoder distribution
$q(z \vert x)$ is shown in Fig.~\ref{fig:probmodel}a. Variational inference
involves a KL regularization of $q(z \vert x)$ towards the prior $p(z)$ and
VAEs choose $q$ such that it can be computed efficiently, e.g. Gaussian
parameterizations \cite{VAE,VAE2} or normalizing flows
\cite{kingma2016improved}.
To increase the flexibility of encoder distributions, \cite{mescheder} uses
an adversarial approach that requires only samples to compute the KL
regularization. Similar to GANs, it involves a two player game. This
time, a classifier has to distinguish between latent codes sampled from the
prior and those sampled from the encoder distribution, and the second player is
the encoder.

However, images are the product of two independent factors, pose $\pose$ and
appearance $\app$. There are both variational \cite{betavae} and adversarial
approaches \cite{hu2018disentangling} that try to discover such disentangled
factors without any additional source of information. But simply assuming
that $\pose$ and $\app$ are different components of $z$, i.e. $z = (\pose,
\app)$ is problematic because the prior $p(z)$ cannot model the individual
contribution of pose and appearance without inductive biases
\cite{challenging,szabo}. The resulting models fall short in comparison to
approaches that can leverage additional information
\cite{cyclevae,kotovenko2019content}.  Thus, to learn a model in which the
generative process is described by $p(x \vert \pose, \app)$ we need
additional information.

\subsection{Pose supervised disentangling}
The common assumption of many recent works on disentangled image generation,
e.g., \cite{PG2,DPIG,VUNet,SIHP,spurr2018cross,esser2018towards}, is the
observability of $\pose$, which is derived from a pretrained model for
keypoint detection. While this representation of $\pose$ works quite well,
it is limited to domains where robust keypoint detectors are available and
sidesteps the learning task of disentangling the two latent factors $\pose,
\app$.  Instead, let us assume for a moment that we can observe samples of
image triplets $(x_1,x_2,x_3)$ with the constraints that \emph{(i)} $x_1$,
$x_2$ have the same appearance, \emph{(ii)} $x_2$, $x_3$ share the same
pose, and \emph{(iii)} $x_1$, $x_3$ have neither pose nor shape in common.
Let 
$\ptriplet(x_1,x_2,x_3)$ denote this unknown joint distribution.
We model each of the three images as being generated by a process of the
form $p(x \vert \pose, \app)$ (which is assumed to be the same for all three
images). Because $x_1, x_2$ share appearance and $x_2,
x_3$ share pose, only four instead of six latent variables are required to
explain how these triplets are generated. Let $\pose, \app$ denote the
shared pose and appearance explaining $x_2$ and let $\pose', \app'$ denote
additional realizations of pose and appearance explaining $x_1$ and $x_3$,
respectively. The marginal distributions underlying $\pose$ and $\pose'$ are
assumed to be the same and so are the marginal distributions of $\app$ and
$\app'$. If, as assumed in \cite{dva,kulkarni2015deep}, we could observe the
complete triplet $(x_1,x_2,x_3)$, a simple inference mechanism would infer
$\app$ from $x_1$ and $\pose$ from $x_3$ as depicted in
Fig.~\ref{fig:probmodel}b.  Unfortunately, the assumption that $x_2, x_3$
share the same pose but not appearance is essentially equivalent to the
assumption that a keypoint estimator is implicitly available. Then pairs
$x_2, x_3$ could be found by comparison of keypoints. Without this
information we have to further reduce assumptions on the data and
essentially train without $x_3$.

\subsection{Disentangling without pose-annotations}
Without access to $x_3$, we must rely on $x_2$ to infer $\pi$ as shown in
Fig.~\ref{fig:probmodel}c. The maximum likelihood objective for $p(x_2 \vert
\pose, \alpha)$ leads to a reconstruction loss, and without constraints on
$\pi$ it encourages a degenerate solution where all information about $x_2$
is encoded in $\pi$ \cite{mathieu,cyclevae}, i.e. $\pi$ also encodes
information about $\alpha$ instead of being independent of it.

\cite{kingmasemi} assumes the availability of labels for $\alpha$ and uses a
conditional variant of the VAE, which results in a KL regularization of
$q(\pi \vert x_2)$ towards a prior and thus a constraint on $\pi$. To
improve image generations with swapped $\alpha$ and $\pi$, \cite{mathieu}
adds an adversarial constraint on generated images. It encourages the
preservation of characteristics of $\alpha$, i.e. it combines the
conditional VAE with a conditional GAN similar to \cite{cvaegan}.
\cite{szabo} also utilize this
GAN constraint but they only require pairs $x_1,x_2$ instead of labels for
$\alpha$, and instead of using the KL term to constrain $\pi$, they
severely reduce its dimensionality. As
pointed out by \cite{cyclevae}, these GAN constraints only encourage the
decoder to ignore information about $\alpha$ in $\pi$ instead of
disentangling $\alpha$ and $\pi$. \cite{cyclevae} proposes a
cycle-consistent VAE which adds a cyclic loss to the VAE objective.
\cite{lorenz2019unsup} directly models $\pi$ as keypoints.
All of these methods rely on the same basic principle for disentanglement: 
Constraining the amount of information in $\pi$. Indeed, the VAE objective
implements a variational approximation of the information bottleneck
\cite{alemi2016deep}. In contrast, we utilize this variational information
bottleneck only to counteract overpowering, an issue that affects the
following methods.

Similar to \cite{mescheder} for variational inference,
\cite{denton2017unsupervised} utilizes an adversarial approach for
disentanglement: A classifier has to predict if a pair $(\pi, \pi')$ was
inferred from two images of the same video sequence or different video
sequences.  \cite{hadad} assumes that $\alpha$ is given in the form of class
labels, and \cite{lample2017fader,attgan} are specialized to images of
faces and assume that $\alpha$ is given in the form of facial attributes but
they utilize the same principle: A classifier has to predict $\alpha$ from
$\pi$. Besides differences in the precise objectives used for the
classifiers, all of these methods implement again an information bottleneck
as in \cite{mine}. Compared to variational approximations of the bottleneck,
they have the advantage that only information about $\alpha$ in $\pi$ is
penalized. However, applications of adversarial approaches have been limited
to synthetic datasets or facial datasets with little to no pose variations.
We show that overpowering prevents their direct application to real world
datasets and show how to turn them into robust methods for disentanglement.
Our approach can recombine pose and appearance of any two images, while
previous models for unsupervised image-to-image translation require seperate
training for each appearance \cite{huang2018multimodal,lee2018diverse} and
cannot transfer to unseen appearances \cite{choi2018stargan}.%
\begin{figure}
        \centering
        \includegraphics[width=\linewidth]{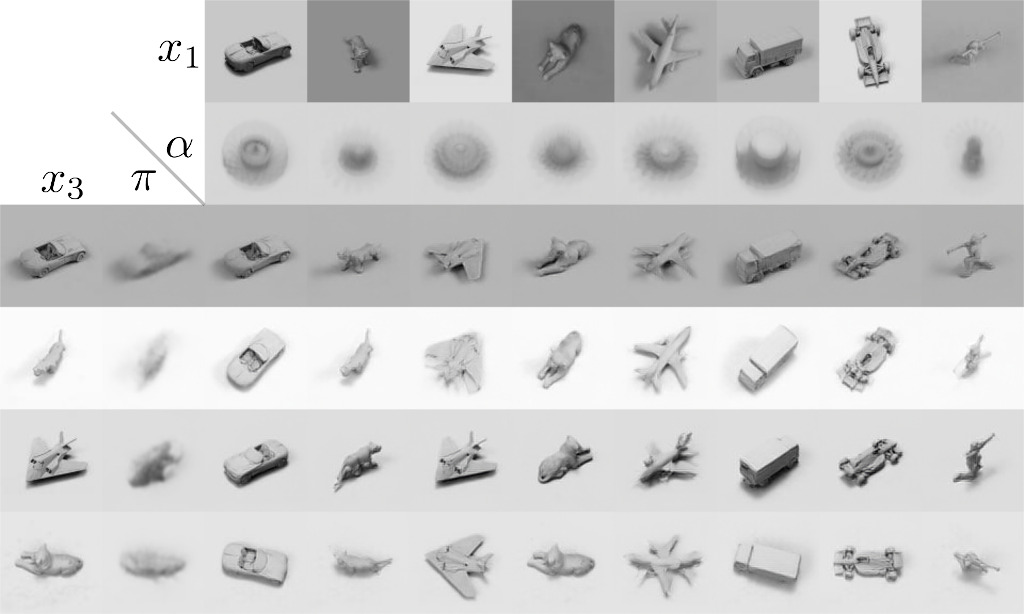}
        \caption{Generated images using examples from the norb
        dataset \citep{norb}. First row: images used as
        appearance target, second: visualizing the
        inferred appearance $\app$ marginalized over all poses. First column:        images used as pose
        target, second: pose marginalized over all appearances. Remaining
        entries: decodings for different combinations of pose and
        appearance.}
        \label{fig:norbmatrix}
\end{figure}

\newcommand{\MI}{I}
\section{Approach}
\subsection{Constrained maximum-likelihood learning}

We want to learn a probabilistic model of images that explains the observed
image $x_2$ in terms of two disentangled representations $\pose, \app$. This
requires a model for the decoder distribution conditioned on the two
representations,
\begin{equation}
  p(x_2 \vert \pose, \app)
\end{equation}
and an encoder model $p(\pose, \app \vert x_1, x_2)$ to infer $\pose$ and
$\app$ from the data.
As shown in Fig.~\ref{fig:probmodel}c, we estimate $\pose$ with an encoder network $E_\pose(x_2)$ from $x_2$ and
$\app$ with an encoder network $E_\app(x_1)$ from $x_1$. 
A decoder network $D(\pose,\app)$ which takes $\pose$ and $\app$ as inputs reconstructs the image according to $p(x_2 \vert \pose, \app)$.
\begin{figure}
        \centering
        \includegraphics[width=\linewidth]{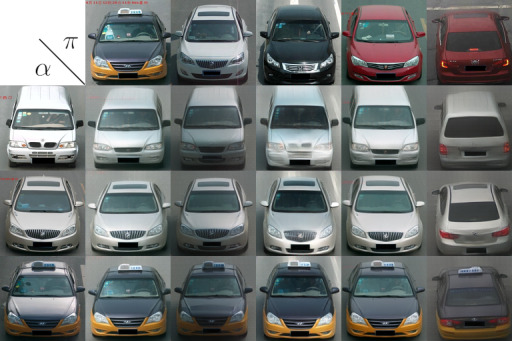}
        \caption{Generated images on the PKU Vehicle ID \cite{vehicles}
        dataset. First row: pose targets. First column: appearance targets.}
        \label{fig:vehiclematrix}
\end{figure}

Learning the weights of these networks depends on a constrained
optimization problem. To ensure that $\pose$ and $\app$ describe the images
well, we maximize the conditional likelihood as formulated in
Eq.~\eqref{objective}, which corresponds to a reconstruction loss. To avoid
a trivial solution where $\pose$ encodes all of the information of $x_2$, we
formulate the disentanglement constraint \eqref{objective2}, such that our
full optimization problem reads
\begin{align}
  \label{objective}
  &\max_{p} \mathbb{E}_{x_1, x_2} \log p(x_2 \vert \pose, \alpha) \\
  \label{objective2}
  &\text{subject to } \MI(\pose, \alpha) \le \epsilon
\end{align}
Here, $\epsilon$ is a small constant and $\MI(\pose, \alpha)$ denotes the mutual information \cite{elements} defined
as
\begin{equation}
  \label{MI}
  \MI(\pose, \app) = \KL(p(\pose, \app) \vert p(\pose) p(\app)).
\end{equation}
Computing \eqref{MI} is difficult \cite{mine} and to derive an algorithm for the
solution of the
optimization problem above, we must resort to
approximations. Subsequently, we first derive two different estimates
on the mutual information. The first one provides an upper bound, but, alas,
it always overestimates it severely. A second estimate is then introduced
which provides accurate estimates. However, to enforce the constraint in
\eqref{objective2}, we require gradients of the estimate and, as we will see,
this enables the encoder to perform an adversarial attack on the estimate, such that it heavily
underestimates the true mutual information. In Sec.~\ref{robustmethod},
we show how to combine both estimates to obtain our
method for robust maximum-likelihood learning under mutual information
constraints. Thereafter, we describe the algorithm used to implement the
method.

\subsection{A variational upper bound on the mutual information}
Ideally, we would like to obtain an upper bound on the MI in
\eqref{MI} to be able to enforce
the constraint \eqref{objective2}.
Because we estimate $\pose$ from $x_2$ and $\app$ from $x_1$, we have the Markov-Chain $\pose \to x_2 \to \alpha$ with
\begin{equation}
  p(\app, x_2, \pose) = p(\app \vert x_2) p(x_2 \vert \pi) p(\pi)
\end{equation}
which implies the data processing inequality \cite{elements}:
\begin{equation}
  \label{dataprocessing}
  I(\pose, \app) \le I(\pose, x_2).
\end{equation}
The right hand side of this inequality can now be easily estimated with a
variational marginal $r(\pose)$ \cite{alemi2016deep}. Indeed, for any density $r$ with
respect to $\pose$ we have the bound
\begin{equation}
  \MI(\pose, \alpha) \le \mathbb{E}_{x_2} \KL(p(\pose \vert x_2) \vert r(\pose)).
\end{equation}
Modeling both $p(\pose \vert x_2)$ and $r(\pose)$ as Gaussian distributions,
we can evaluate the right hand side analytically. Unfortunately, this bound is
too loose for our purposes. The condition $\KL(p(\pose \vert x_2) \vert
r(\pose)) = 0$ implies $I(\pose, x_2) = 0$ and thus $\pose$ would be
completely uninformative.

\subsection{Fine-grained estimation of mutual information}
A different estimate of mutual information can be obtained with the help of
density estimation \cite{mescheder, mine}. The KL-divergence of two
densities is
closely related to the associated classification problem:
Let $T(\pose, \alpha)$ be a classifier that
maps a pair $(\pose, \alpha)$ to a real number which represents
the log probability that the pair is a sample from the joint distribution
$p(\pose, \alpha)$. Denote by $\sigma(t) = (1+e^{-t})^{-1}$ the sigmoid function. The maximum
likelihood objective for this classification task reads
\begin{align}
  \label{classification}
  \max_T\;
  &\mathbb{E}_{(\pose, \alpha) \sim p(\pose, \alpha)}
  \log \sigma(T(\pose, \alpha)) + \\
  &\mathbb{E}_{\pose \sim p(\pose), \app \sim p(\app)}
  \log (1 - \sigma(T(\pose, \app))).
\end{align}
The optimal solution $T^*$ of this problem satisfies
\begin{equation}
  \MI(\pose, \app) = \mathbb{E}_{(\pose, \alpha)
  \sim p(\pose, \alpha)} T^*(\pose, \alpha).
\end{equation}
When $T$ is
implemented as a neural network, we obtain a differentiable estimate of
$\MI(\pose, \app)$ which can be used to enforce the desired constraint during
learning of $q(\pose \vert x_2)$. For a given classifier $T$ we write
$\MI_T=\MI_T(\pose, \app) = \mathbb{E}_{\pose,\app \sim p(\pose, \app)} T(\pose,
\app)$ for the resulting estimate.
\begin{figure}
        \centering
        \includegraphics[width=\linewidth]{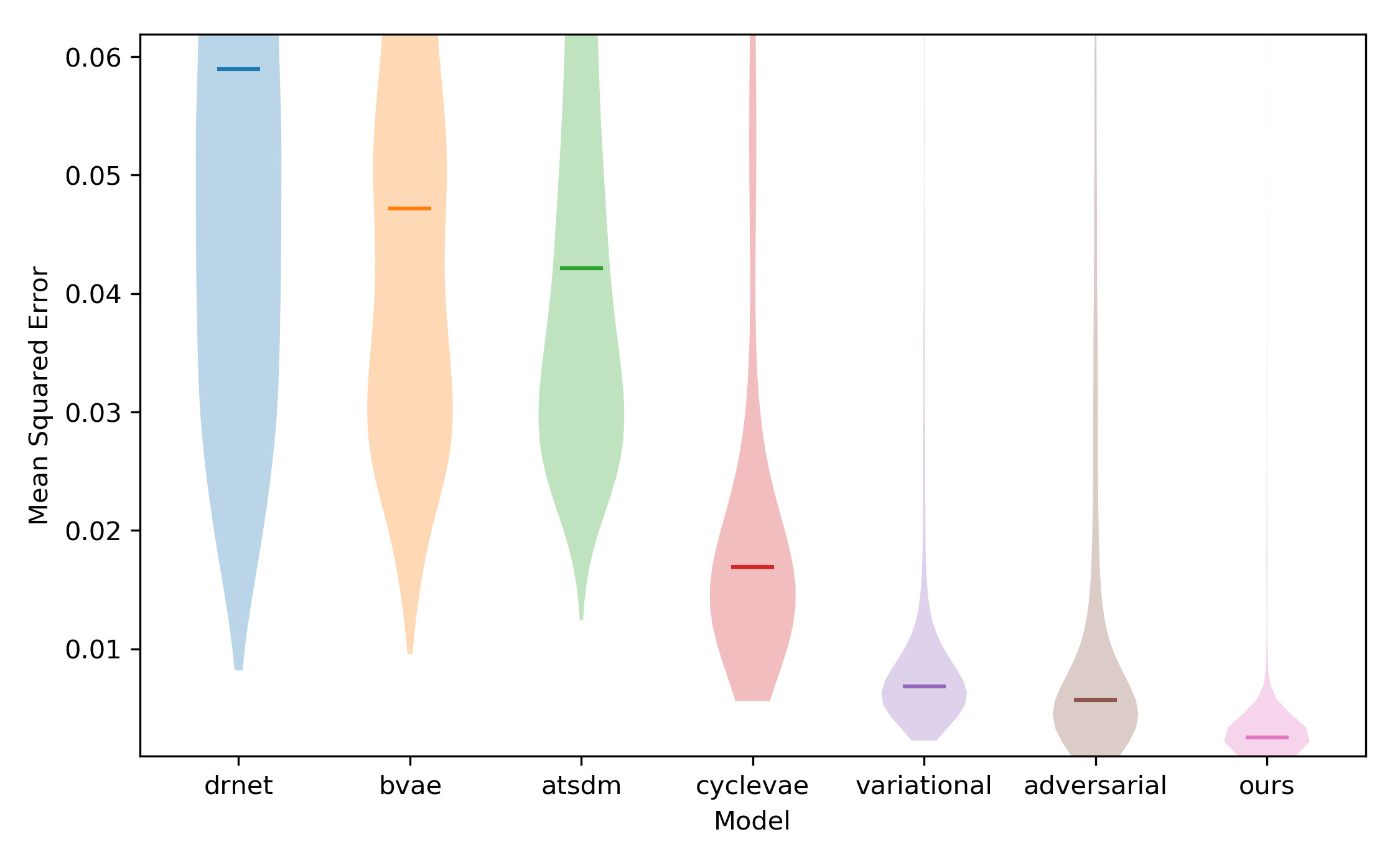}
        \caption{Distribution of average reconstruction error on the sprites
        dataset \cite{dva}. We evaluate against the provided ground
        truth.}
        \label{fig:errors}
\end{figure}

\subsection{Robust combination of variational and adversarial estimation}
\label{robustmethod}
If we replace the constraint $\MI(\pose, \app) \le \epsilon$ in
\eqref{objective2} with a constraint on the
estimate $\MI_T(\pose, \app) \le \epsilon$, we observe a new type of adversarial attack:
The encoder is able to overpower the classifier $T$: it can learn a distribution $p(\pose \vert \app)$ such that
$T$ cannot differentiate pairs $(\pose, \app)$ sampled from the joint
from those sampled from the marginals. However, a seperately trained
classifier $T'$, whose gradients are not provided to the encoder, can
still classify them (see Fig.~\ref{tab:abl}). In other words, in an
adversarial setting we consistently observe the situation $I_T(\pose, \app)
\ll I(\pose, \app)$, i.e., we underestimate the mutual information between
$\pose$ and $\app$. To obtain a guaranteed upper bound on the mutual
information, we must
utilize the variational upper bound. As we have seen before, we must be
careful to enforce not too strict bounds on it. Thus, we formulate our new objective as
\begin{align}
  \label{newobjective}
  &\max_{p} \mathbb{E}_{x_1, x_2} \log p(x_2 \vert \pose, \alpha) \\
  \label{miconstraint}
  &\text{subject to } \MI_T(\pose, \alpha) \le \epsilon \\
  \label{klconstraint}
  & \phantom{subject to }\KL(p(\pose \vert x_2) \vert r(\pose)) \le C,
\end{align}
where $C$ has to be adaptively estimated based on the detection of adversarial
attacks of the encoder
against $T$. The main idea is to compare the classification performance of $T$ against an
independently trained classifier. If there is a large performance gap,
we cannot rely on the estimate of $T$ (it has been overpowered) and must decrease $C$.
The next section describes the approach.
\subsection{Robust disentanglement despite encoder overpowering}

\label{sec:inference}
To obtain a training signal for our networks, we must transform problem
\eqref{newobjective} into an unconstrained problem which can be optimized by
gradient ascent.
Let us
first consider the constraint \eqref{klconstraint} on the $\KL$ term. For a
given $C$, there exists a Lagrange multiplier $\gamma \ge 0$ such that the problem
can be written equivalently as
\begin{align}
  &\max_{p} \mathbb{E}_{x_1, x_2} \log p(x_2 \vert \pose, \alpha) - \gamma
  \KL(p(\pose \vert x_2) \vert r(\pose)) \\
  \label{remainingconstraint}
  &\text{subject to } \MI_T(\pose, \alpha) \le \epsilon .
\end{align}
Thus, we can directly estimate $\gamma$ instead of $C$.
Ideally, $\gamma$ should be very small and only active in situations
were $I_T(\pose, \app)$ underestimates the mutual information. To achieve
this, we train a second classifier $T'$ based on the same objective
\eqref{classification}. It is crucial that its estimate $I_{T'}(\pose, \alpha)$ is never directly provided as a signal to the encoder. We merely compare the
estimates of $T$ and $T'$ and if $I_{T} \ll I_{T'}$, we
increase $\gamma$. Hence, we update $\gamma$ in each optimization step based
on the proportional gain $I_{T'} - I_{T}$ and bias it towards zero with a
small constant $b_\gamma$
\begin{equation}
  \label{adagamma}
  \gamma_{t+1} = \max\{0, \gamma_t + l_\gamma (I_{T'} - I_{T} - b_\gamma)\},
\end{equation}
where $l_\gamma$ can be considered the learning rate of $\gamma$.

\begin{figure}
        \centering
        \includegraphics[width=\linewidth]{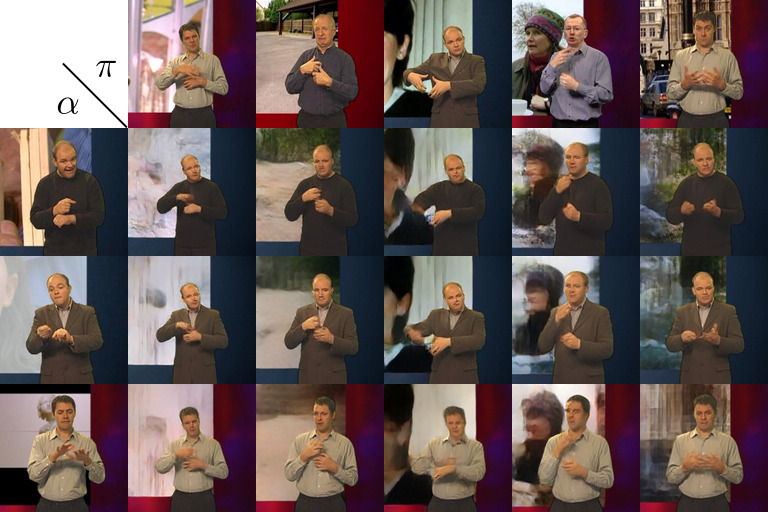}
        \caption{Generated images on the BBC Pose dataset \cite{bbcpose}
        dataset. First row:
        pose target. First column: appearance target. An animated version can be found in the supplementary.}
        \label{fig:bbcmatrix}
\end{figure}

For the remaining constraint \eqref{remainingconstraint} on
$I_T(\pose, \app)$, we utilize an Augmented Lagrangian Approach
\cite{nocedal}. After switching from maximization to minimization, the
complete unconstrained loss function $\mathcal{L}$ for training the network is
\begin{equation}
   \Lall = \Lrec + \Lvb + \Lmi
  ,
\end{equation}
where $\Lrec$ is the reconstruction loss given by the negative likelihood
\begin{equation}
  \Lrec = - \mathbb{E}_{x_1, x_2} \log p(x_2 \vert \pose, \alpha),
\end{equation}
$\Lvb$ the penalty associated with the variational upper bound
\begin{equation}
  \label{lossvb}
  \Lvb = \gamma \KL(p(\pose \vert x_2) \vert r(\pose)),
\end{equation}
and $\Lmi$ the loss used to enforce the the constraint
\eqref{remainingconstraint} based on an estimated Lagrange multiplier
$\lambda \ge 0$ and a penalty parameter $\mu > 0$
\begin{equation}
  \Lmi =
  \begin{cases}
    \lambda (\MI_T - \epsilon) + \frac{\mu}{2} (\MI_T - \epsilon)^2 \quad
    &\text{if } \MI_T - \epsilon \ge - \frac{\lambda}{\mu}\\
    -\frac{\lambda^2}{2\mu} \quad &\text{else}
  \end{cases}.
\end{equation}
The update rule for $\lambda$ is
\begin{equation}
  \label{adalambda}
  \lambda_{t+1} = \max\{0, \lambda_t + \mu (\MI_T - \epsilon)\}.
\end{equation}

Fig.~\ref{fig:model} outlines our network architecture during training and inference. We perform the optimization over mini-batches and
alternate between the training of the classifiers $T$ and $T'$ (according to
the objective defined in \eqref{classification}), and the training of the
generative model. The loss for the networks $D$ and $E_\app$ is given
by $\Lrec$ and $E_\pose$ is optimized with respect to the full loss $\Lall$.
After each step, $\gamma$ and $\lambda$ are updated according to
\eqref{adagamma} and \eqref{adalambda}, respectively.

\begin{figure}
        \centering
        \includegraphics[width=\linewidth]{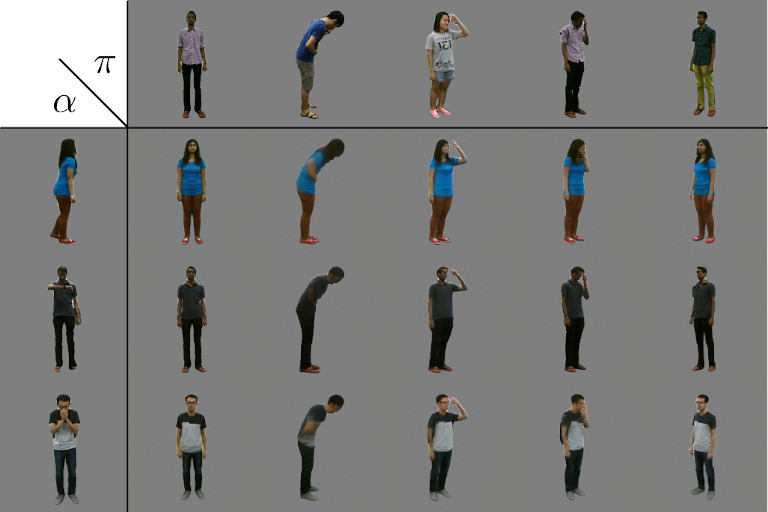}
        \caption{Generated images on the NTU dataset \cite{ntu}. First row:
        pose target. First column: appearance target. An animated version can be found in the supplementary.}
        \label{fig:ntumatrix}
\end{figure}

\newcommand{\tabsection}[2]{%
    \cmidrule{2-7}
    &
    \multicolumn{6}{c}{\cellcolor[gray]{0.9}
        \textbf{#1} #2
    } \\
    \cmidrule{2-7}
}

\newcommand{\colsection}[2]{%
        \multirow{#1}{*}{\rotatebox[origin=c]{90}{#2}}
}

\newcommand{\coltabsection}[2]{}

\begin{table*}
\centering
\begin{tabular}{ccccccc}
\toprule
&
&
& 
\multicolumn{2}{c}{Market-1501 \citep{market}} & 
\multicolumn{2}{c}{DeepFashion \citep{deepfashion1,deepfashion2}} \\
        \cmidrule(lr){4-5} 
        \cmidrule(lr){6-7} %
&
&
Model &
Reconstruction &
Transfer &
Reconstruction &
Transfer \\
  \vspace{-1em}
\colsection{6}{Appearance} \\

\tabsection{reID}{mAP $[\si{\percent}]$
                  - \emph{bigger is better}}

& \multirow{1}{*}{sup.}
&
VUNet \citep{VUNet}
& \num{25.3}  %
& \num{19.9}  %
& \num{21.3}  %
& \num{14.6} \\  %
\cmidrule(lr){2-7}
& \multirow{2}{*}{unsup.}
&
adversarial
& \num{34.8}  %
& \num{6.6}  %
& \num{37.4}  %
& \num{10.5} \\  %

&
&
ours 
& \num{30.2}  %
& \num{25.4}  %
& \num{52.9}  %
& \num{47.1} \\  %

\midrule
\tabsection{rePose}{$[\si{\percent}\,\text{of image width}]$
                    - \emph{smaller is better}}
\colsection{0}{Pose}
& \multirow{1}{*}{sup.}
&
VUNet \citep{VUNet}
& \num{17.9 +- 9.4}  %
& \num{17.5 +- 9.4}  %
& \num{1.5 +- 3.1}  %
& \num{1.9 +- 4.0} \\  %
\cmidrule(lr){2-7}
& \multirow{2}{*}{unsup.}
&
adversarial
& \num{24.6 +- 10.3}  %
& \num{25.3 +- 10.3}  %
& \num{6.2 +- 7.3}  %
& \num{7.4 +- 8.1} \\  %

&
&
ours
& \num{23.9 +- 9.9}  %
& \num{24.7 +- 9.9}  %
& \num{5.5 +- 6.8}  %
& \num{6.8 +- 7.6} \\  %

\bottomrule
\end{tabular}
\caption{Evaluating how well the generated image preserves \emph{(i)}
appearence or \emph{(ii)} pose. For
\emph{(i)} we compare input $\xapp$ and output $\xout$ of our approach using
a standard encoder for person reidentification \cite{tripreid} using
retrieval performance (mAP). Conservation of pose \emph{(ii)} is measured by
comparing the results of keypoint detector \citep{opose} of $\xpose$ and
$\xout$. Note that \citep{VUNet} uses keypoint annotations.}%
\label{tab:res}
\end{table*}

\section{Experiments}
\label{sec:experiments}

\subsection{Comparison to state-of-the-art}

In Fig.~\ref{fig:errors}, we compare our method to \cite{cyclevae}
(cyclevae), the state-of-the-art among the variational approaches, and to
\cite{hadad} (atsdm), the state-of-the-art among the adversarial approaches.
In addition to our full model (ours), we also include
\cite{denton2017unsupervised} (drnet), a version of our model
that utilizes the objective of \cite{betavae} (bvae), a version without
$\Lvb$ (adversarial) and a version without $\Lmi$ (variational), where the update of
$\gamma$ from \eqref{adagamma} is replaced by
\begin{equation}
  \label{adagamma2}
  \gamma_{t+1} = \max\{0, \gamma_t + l_\gamma (I_{T'} - \epsilon)\},
\end{equation}
to estimate the required $\gamma$ to achieve the MI constraint
\eqref{objective2}.

Because it is difficult to obtain ground truth for triplets $(x_1, x_2,
x_3)$ on real data, we resort to the synthetic sprites dataset \cite{dva}
to compare the methods. It contains 672 different video game characters,
each depicted in a wide variety of poses. For training, we only utilize
pairs of images $(x_1, x_2)$ belonging to the same character. To measure the
performance of the different approaches, we calculate the mean squared error
between images $\tilde{x}_2$ generated from inputs $x_1$ and $x_3$, and the
corresponding ground truth $x_2$. We randomly select 8000 triplets $(x_1,
x_2, x_3)$ from the test set and report the error distribution in
Fig.~\ref{fig:errors}.

\subsection{Visualization of encodings}
\label{visualizations}
To better understand the information encoded by $\pose$ and $\app$,
we visualize these representations. Because $\pose$ does not contain information
about the appearance, this corresponds to a marginalization of images
depicting a given pose over all appearances. Similarly, $\app$ yields a
marginalization for a given appearance over all poses. We show examples of
these visualizations in Fig.~\ref{fig:rmatrix},
Tab.~\ref{tab:abl},
and Fig.~\ref{fig:norbmatrix}. This synthesis is
performed independently from the training of our model with the sole purpose
of interpretability and visualization. For this, a decoder network is
trained to reconstruct images from only one of the factors.
In Tab.~\ref{tab:abl}, these visualizations demonstrate that $I_{T'}$
estimates entanglement correctly. In that figure, a) is our model without
$\Lmi$ and the update of $\gamma$ replaced by \eqref{adagamma2}. b) is our
model without $\Lvb$, c) with $b_\gamma = 0$, d) with $\gamma = 1$ fixed and
e) is our full model.

\subsection{Shared representations across object categories}
The previous dataset contained a single category of objects, namely video
game characters. In this setting, a common pose representation is relatively
easily defined in terms of a skeleton. It is considerably more difficult to
find a representation of pose that works across different object categories
which do not share a common shape. To evaluate our model in
this situation, we utilize the NORB dataset \cite{norb}, which contains
images of 50 toys belonging to 5 different object categories. Each instance
is depicted under a wide variety of camera views and lighting conditions.
For this experiment, we consider camera views and lighting conditions
to be represented by $\pose$.
In
Fig.~\ref{fig:norbmatrix} we can see that our model successfully finds two
representations that can be combined \emph{across} different object
categories. Note that our model was never trained on a pair of images
depicting instances of different categories.

\subsection{Evaluation on Human Datasets}

\newcommand{\emb}{\ensuremath{\tilde{e}}}
\newcommand{\eapp}{\ensuremath{\emb_{\idxapp}}\xspace}
\newcommand{\eout}{\ensuremath{\emb_{\idxout}}\xspace}

\newcommand{\centr}{\ensuremath{\mu}}
\newcommand{\capp}{\ensuremath{\centr_{\idxapp}}\xspace}
\newcommand{\cout}{\ensuremath{\centr_{\idxout}}\xspace}

\newcommand{\dist}{\ensuremath{d}\xspace}

\newcommand{\kp}{\ensuremath{k}}
\newcommand{\conf}{\ensuremath{c}}
\newcommand{\kppose}{\ensuremath{\kp_{\idxpose}}\xspace}
\newcommand{\kpout}{\ensuremath{\kp_{\idxout}}\xspace}

In Tab.~\ref{tab:res}, we evaluate our approach on natural images of people, which have been
the subject of recent models for disentangled image generation \cite{VUNet}.
Besides qualitative evaluations, we employ two quantitative measures to validate how much of the pose
and appearance are being preserved in the generated output:
(i) Since ground-truth triplets are not available for these datasets, we require
a metric that captures
similarity in appearance while being invariant to changes in pose. Such a
measure can be obtained from a person reidentification
model \cite{tripreid}, which can identify the same person despite
differences in pose.
Using the evaluation protocol of \cite{tripreid} we report the
mean average precision (mAP) of re-identifying generated images under
``reID mAP'' in Tab.~\ref{tab:res}.
(ii) To measure how well our approach retains pose
we employ Openpose \citep{opose} to obtain keypoint estimates.
We extract keypoints from the pose input image
\xpose and the output \xout and compute the euclidean distance between the
estimated keypoints in both images.
As above, we include an ablation (\emph{adversarial}) without $\Lvb$.

\section{Conclusion}
We have shown how an additional classifier, whose gradients are not used
directly to train the encoder,
prevents encoder overpowering. This enables robust learning of disentangled
representations of pose and appearance without requiring a prior on pose
configurations, pose annotations or keypoint detectors. Our
approach can be readily applied on a wide variety of real-world datasets.
\blfootnote{This work has been funded by the German Research Foundation
(DFG) - 371923335; 421703927 and a hardware donation from NVIDIA.}

\FloatBarrier
\clearpage

{\small
\bibliographystyle{ieee_fullname}
\bibliography{ms}
}

\appendix
\setcounter{page}{1}
\onecolumn

\begin{center}
	\textbf{
	\Large Supplementary materials for\\
	\Large Robust Disentanglement on Real-World Datasets without
  Pose-Annotations
	}
\end{center}

We provide additional results obtained by our method in
Sec.~\ref{imga} and implementation details in Sec.~\ref{impl}.

\section{Synthesis Results}
\label{imga}

\begin{figure}
	\begin{center}
		\includegraphics[width=0.9\textwidth]{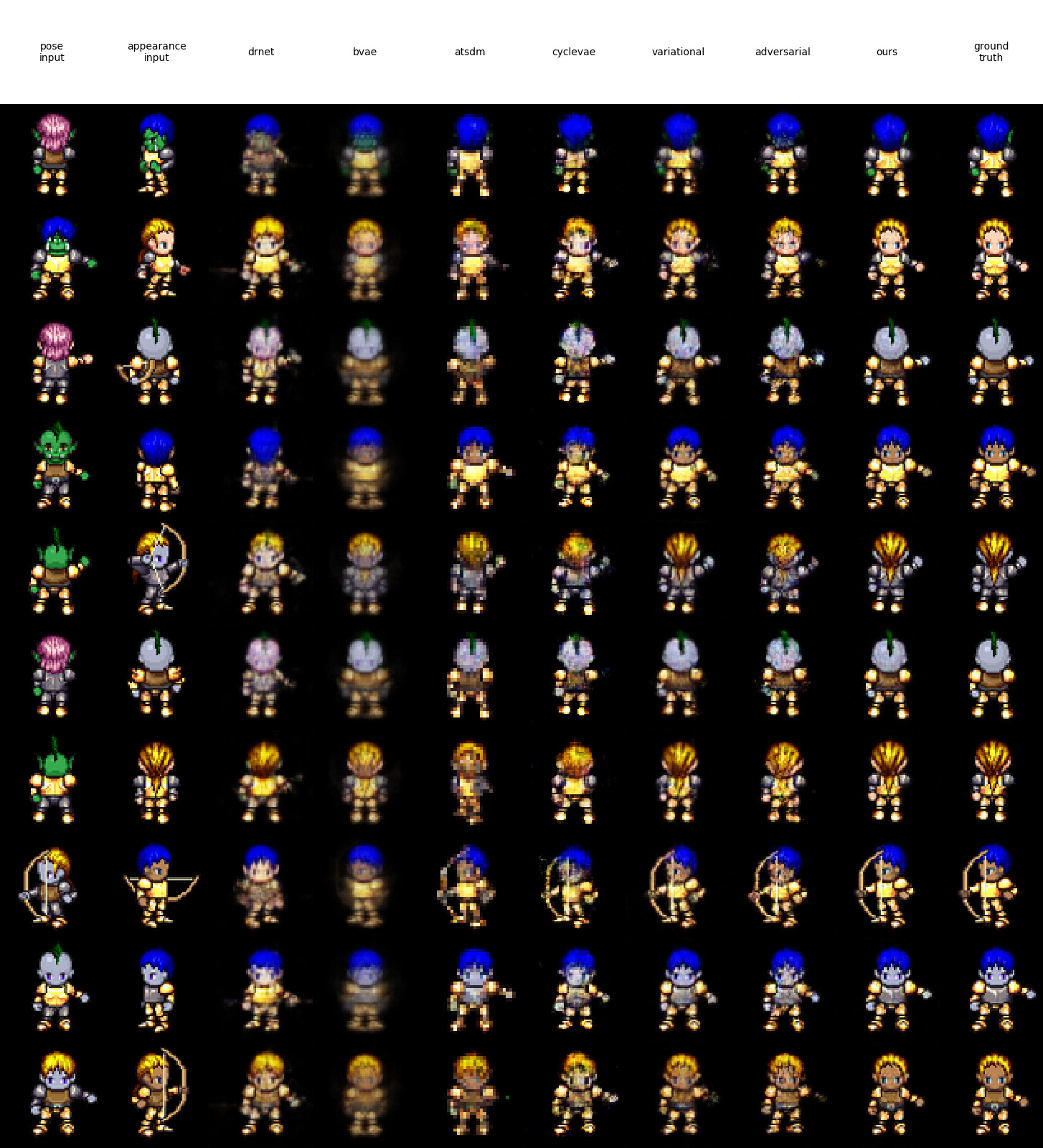}
	\end{center}
  \caption{Qualitative results for the comparison in Fig.~\ref{fig:errors}.}
	\label{fig:supp_qualitative}
\end{figure}

In Fig.~\ref{fig:supp_qualitative} we show some qualitative results of the
comparison in Fig.~\ref{fig:errors}.

\subsection{Object Image Generation}
\begin{figure}
	\begin{center}
		\includegraphics[height=0.9\textheight]{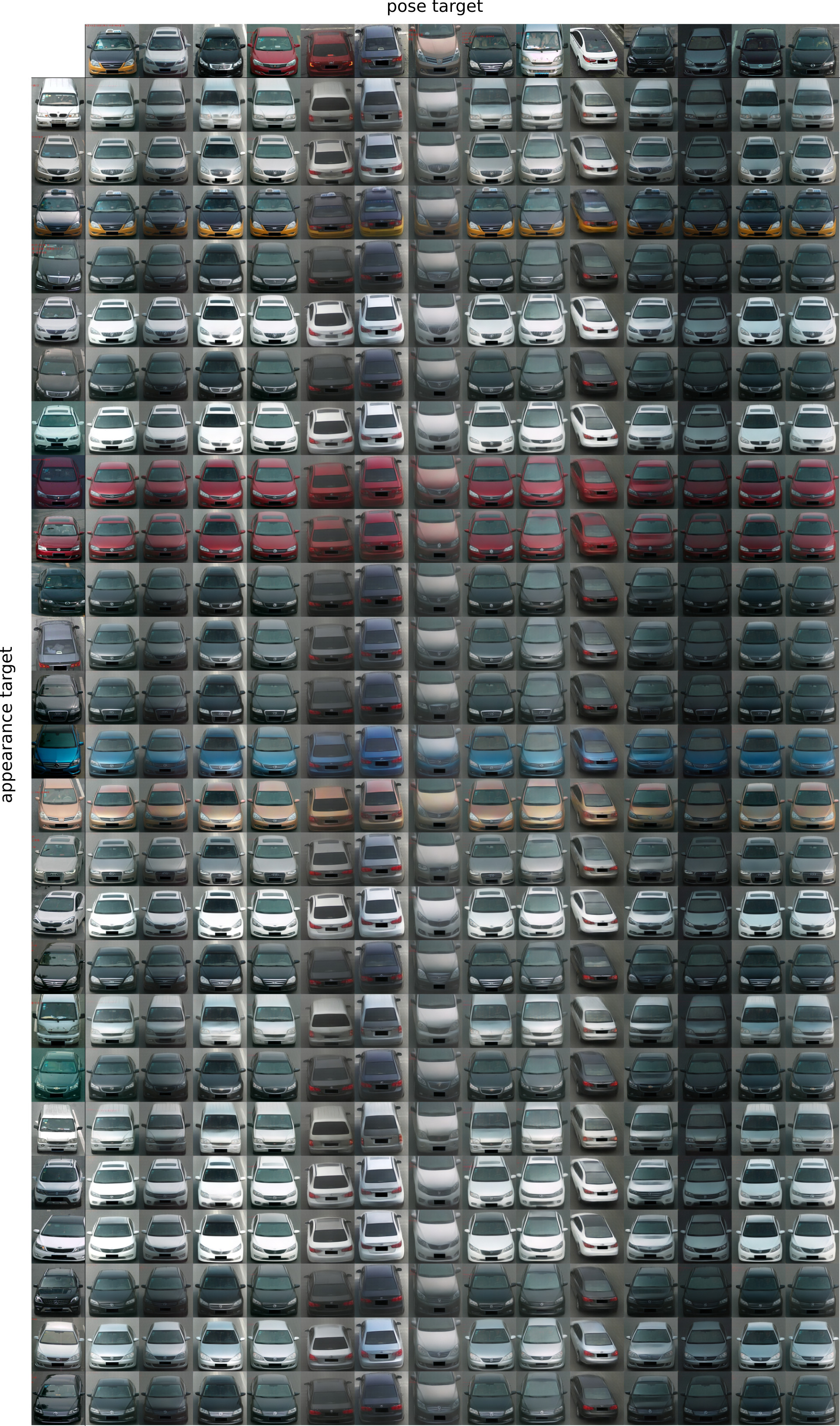}
	\end{center}
  \caption{Retargeting on the PKU Vehicle ID \cite{vehicles} dataset.
  Without changes in the architecture, our method handles both rigid
  objects, as seen here, as well as deformable and articulated objects such
  as humans.}
	\label{fig:supp_vi_matrix}
\end{figure}

Since our method does not rely on the existence of pose estimates, it is
applicable to a wide range of objects. In Fig.~\ref{fig:supp_vi_matrix} we show
additional results of our method obtained on a vehicle surveillance dataset
\cite{vehicles}.

\begin{figure}
	\begin{center}
		\includegraphics[height=0.9\textheight]{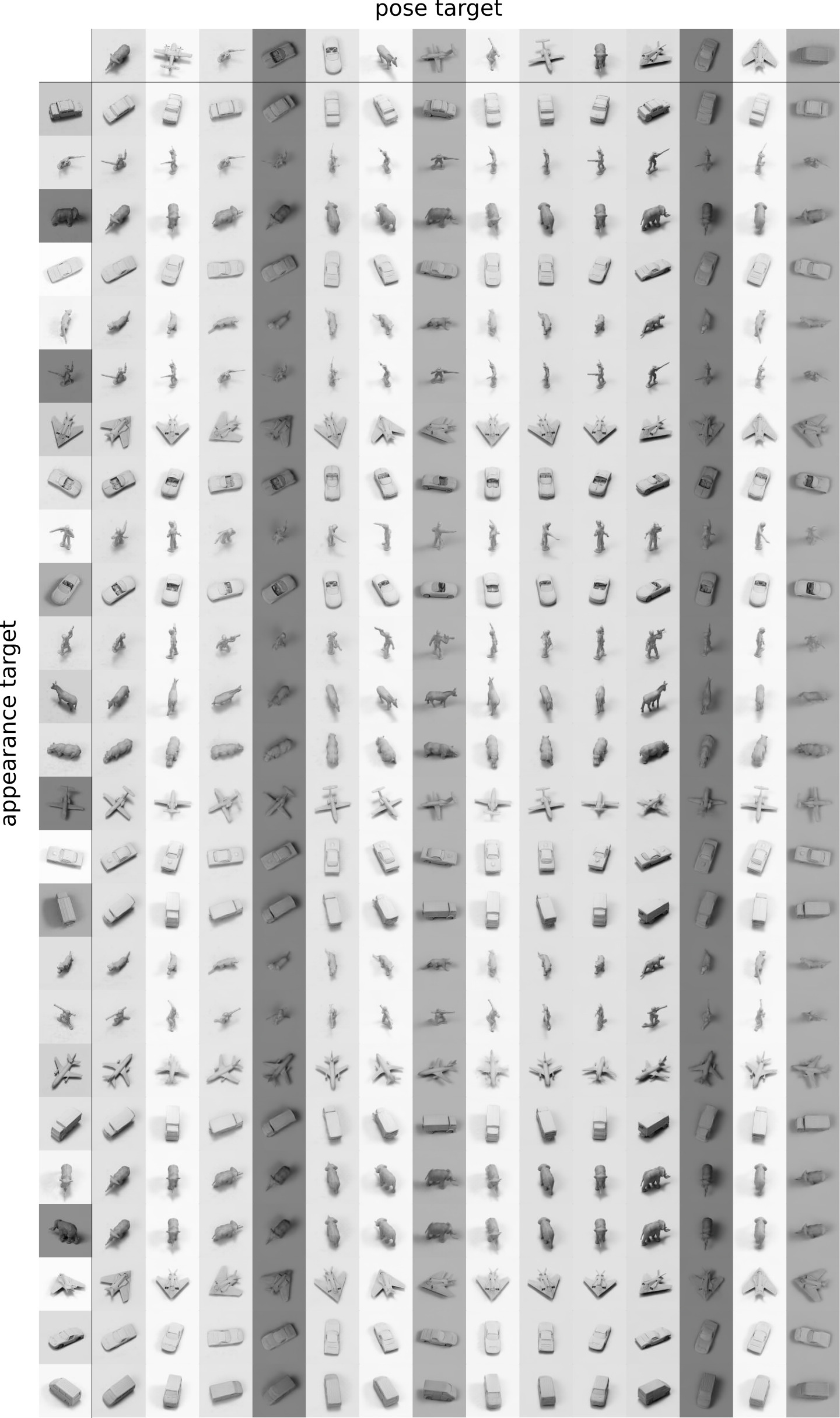}
	\end{center}
  \caption{Retargeting on the Norb dataset \cite{norb}. Our method
  successfully finds a shared representation for pose, which can be used to
  retarget poses across different object categories. An animated version can
  be found in \texttt{norb.avi}, where the target pose is rotated and after each
  full turn, the elevation is increased.}
	\label{fig:supp_norb_matrix}
\end{figure}

Fig.~\ref{fig:supp_norb_matrix} shows that our method is not limited to a single
object category. Indeed, it can learn shared pose representations across different
categories, such as those contained in the norb dataset \cite{norb}.

\subsection{Person Image Generation}
\label{imgb}
In Tab.~\ref{vucompare} we provide a qualitative comparison to \cite{VUNet}
which highlights the benefits of not requiring keypoint estimates even for
domains where keypoint estimators are available. Our approach does not
suffer from pose estimation errors and, compared to keypoints, our pose
representation is better disentangled from appearance.

\begin{figure}
	\begin{center}
		\includegraphics[height=0.9\textheight]{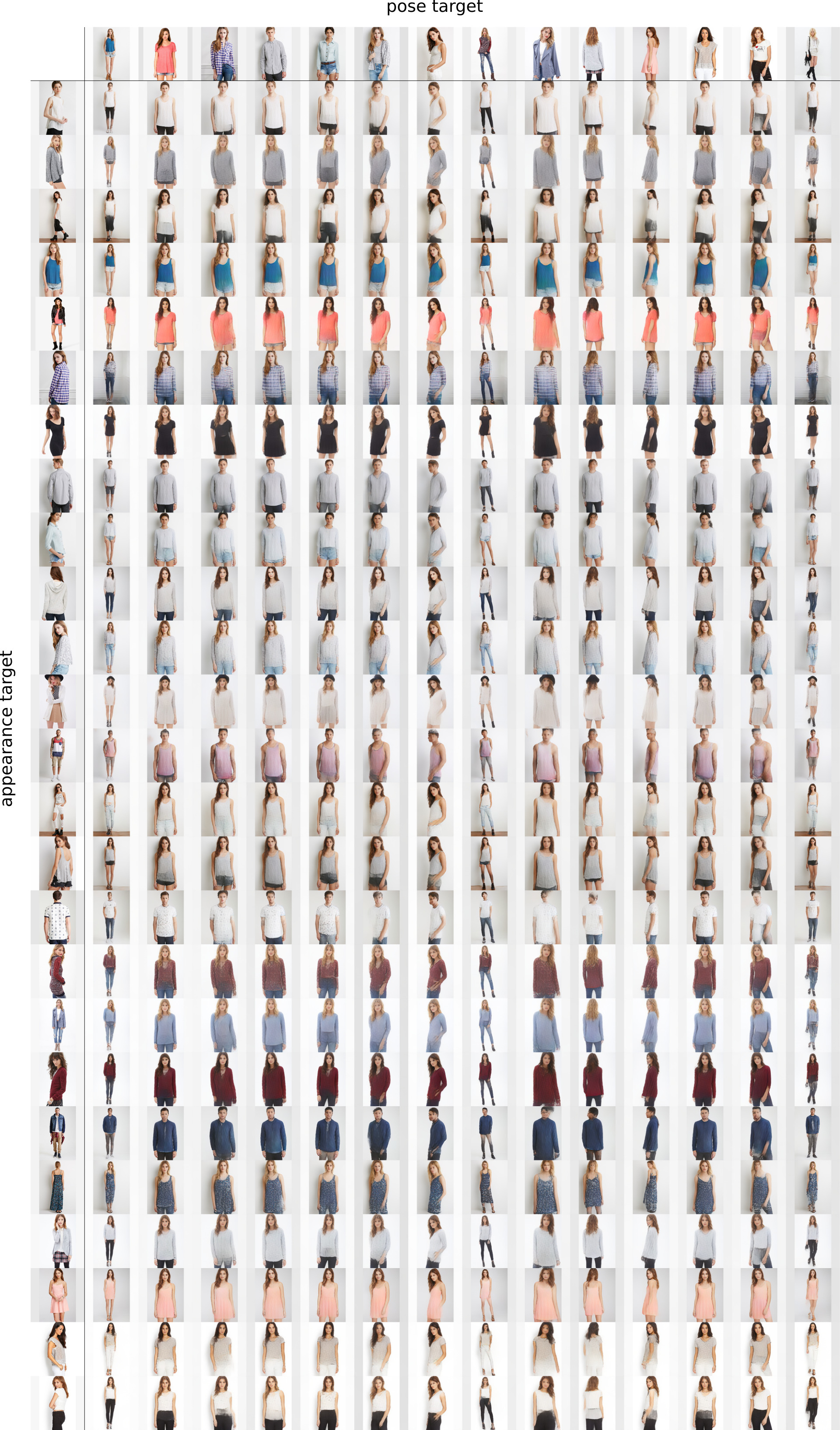}
	\end{center}
  \caption{Retargeting on the DeepFashion dataset
  \cite{deepfashion1,deepfashion2}. Note how our method is able to retarget
  appearances to a wide range of poses, including a change from half-body to
  full-body views. Similiarly, large appearance changes (e.g. changes in
  gender) are possible while
  retaining the pose. Training requires only pairs of images containing the
  same appearance.}
	\label{fig:supp_df_matrix}
\end{figure}

\begin{figure}
	\begin{center}
		\includegraphics[width=0.9\textwidth]{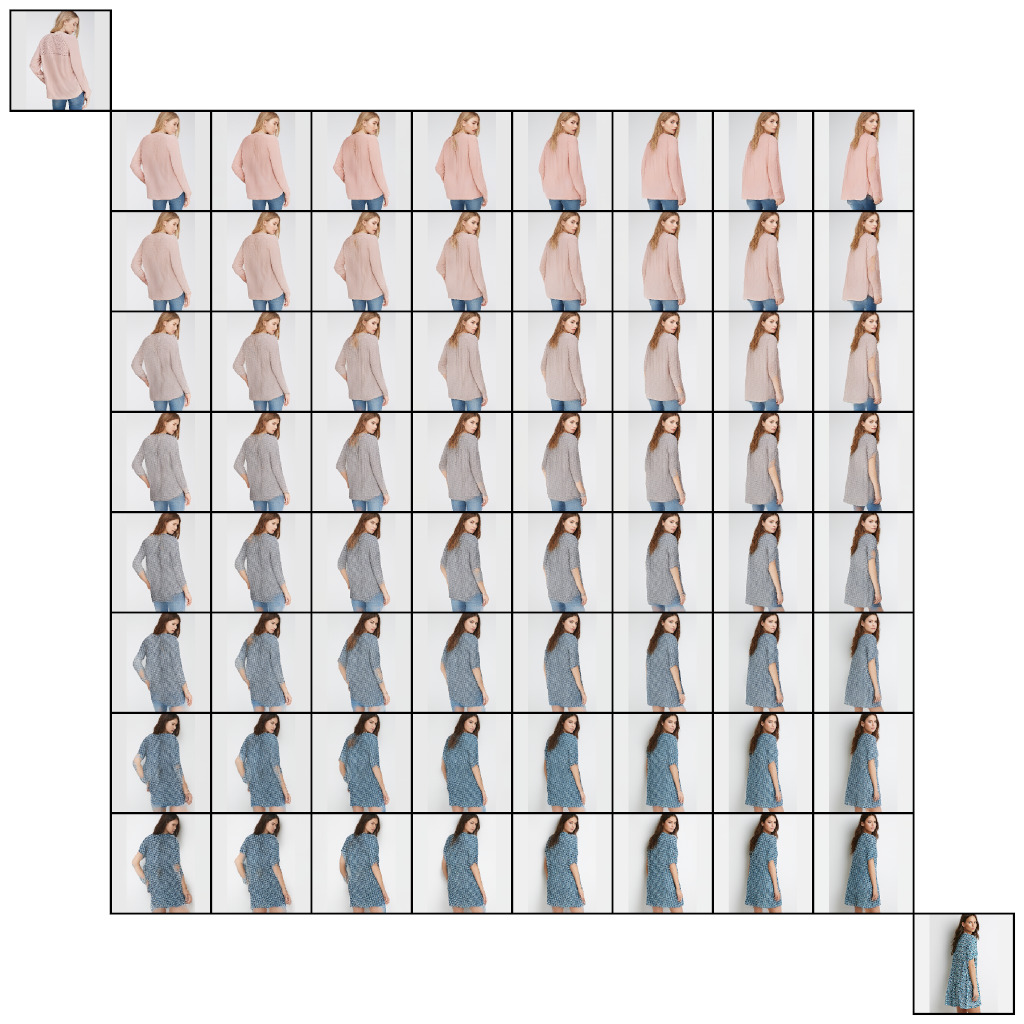}
	\end{center}
  \caption{Interpolating between appearance (vertical direction) and pose
  (horizontal direction).}
	\label{fig:supp_interpolate}
\end{figure}

\begin{figure}
	\begin{center}
		\includegraphics[width=0.9\textwidth]{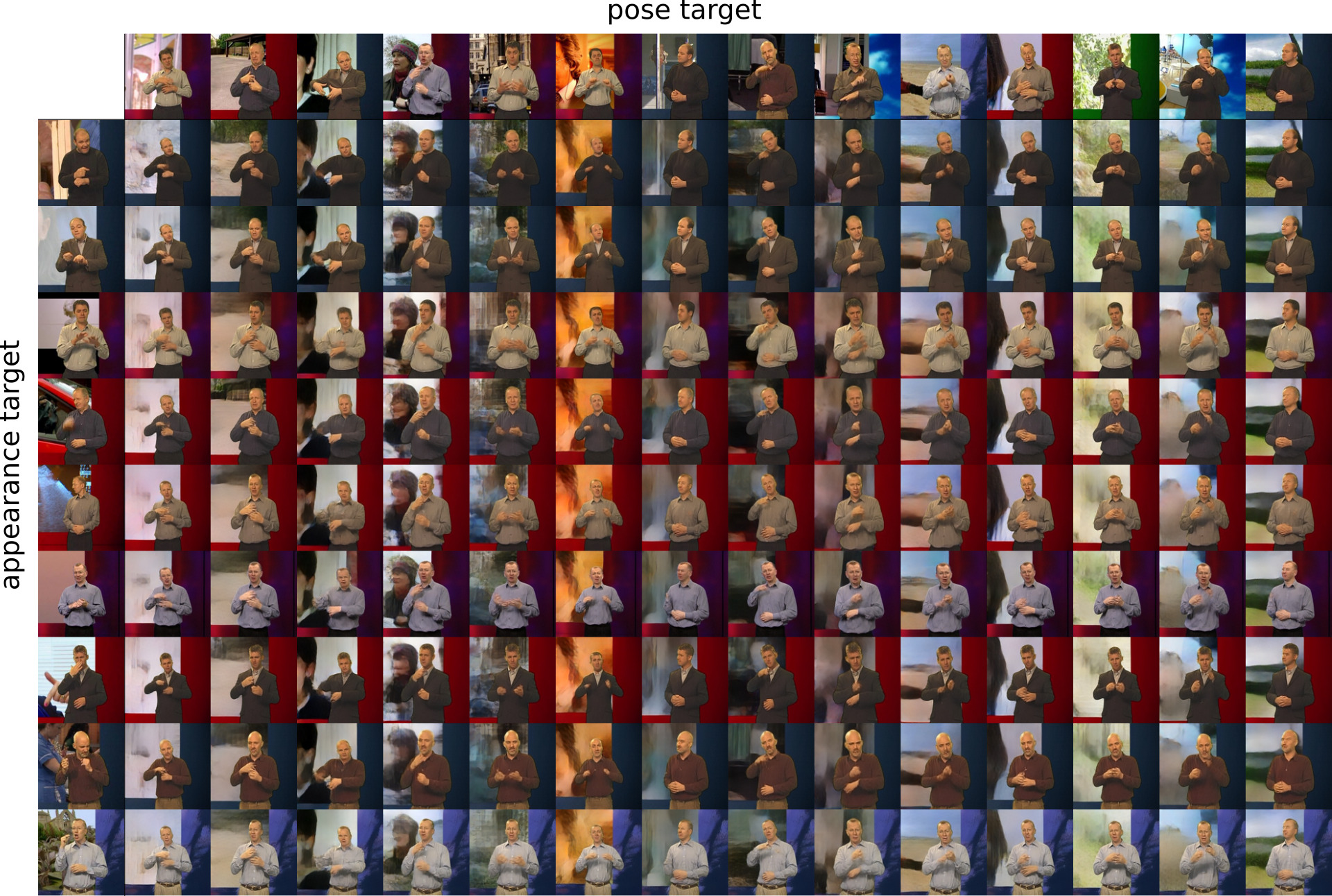}
	\end{center}
  \caption{Retargeting on the BBC Pose dataset \cite{bbcpose}. No
  annotations are required. Our method can utilize different frames from a
  video to learn the transfer task. An animated version can be found in
  \texttt{bbc.avi}.}
	\label{fig:supp_bbc_matrix}
\end{figure}

\begin{figure}
	\begin{center}
		\includegraphics[width=0.9\textwidth]{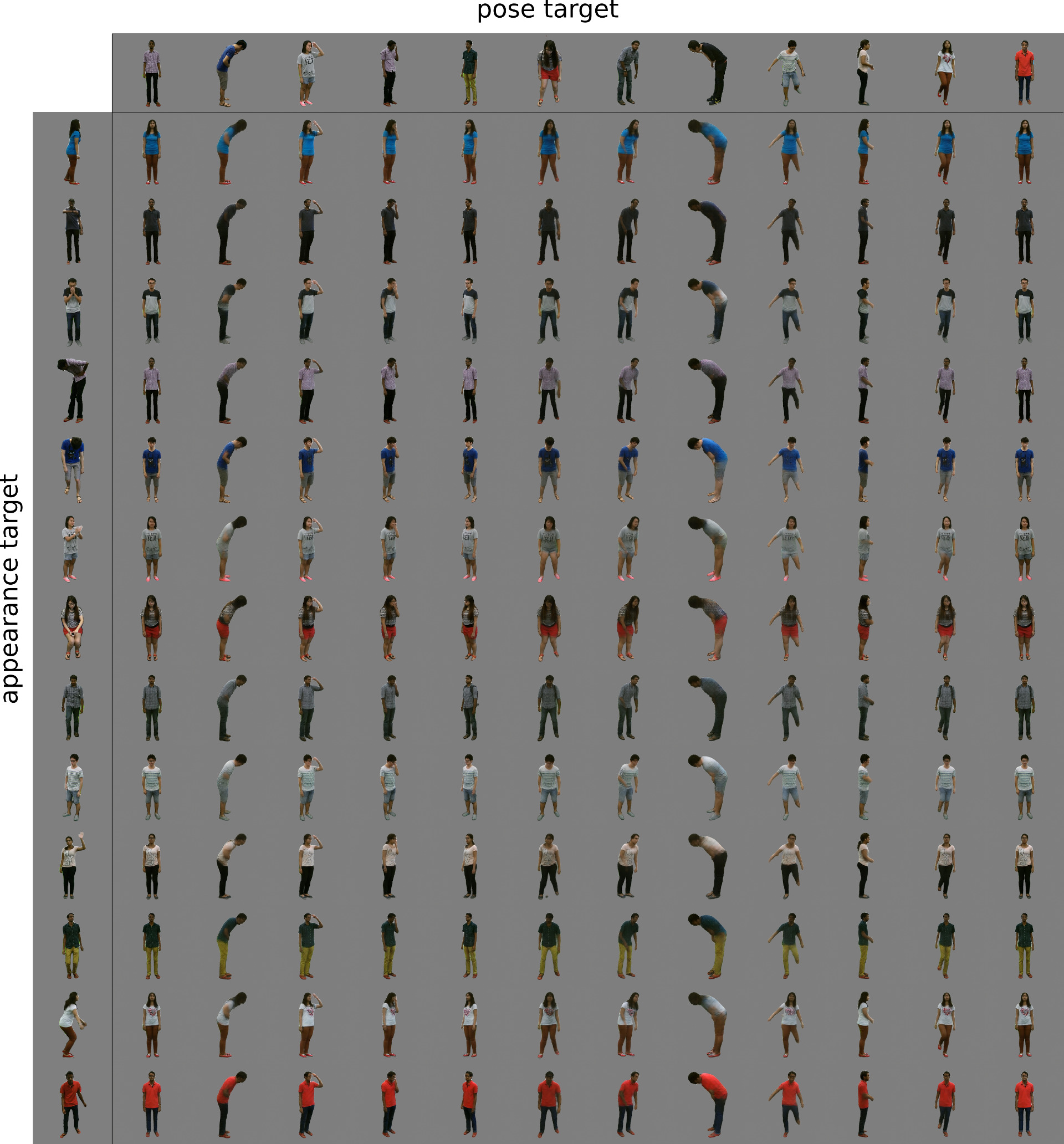}
	\end{center}
  \caption{Retargeting on the NTU dataset \cite{ntu}. Again, our model
  directly learns to disentangle pose and appearance using only video data
  without requiring additional annotations. An animated version can be found
  in \texttt{ntu.avi}.}
	\label{fig:supp_ntu_matrix}
\end{figure}

We show additional results for Person Image Generation in
Fig.~\ref{fig:supp_df_matrix}, Fig.~\ref{fig:supp_bbc_matrix} and
Fig.~\ref{fig:supp_ntu_matrix}. Note that our method learns an appearance
invariant representation of pose across a wide range of poses and viewpoints
(Fig.~\ref{fig:supp_df_matrix}). Fig.~\ref{fig:supp_bbc_matrix} shows that
it can handle fine grained pose representations such as those required for
hands. Fig.~\ref{fig:supp_ntu_matrix} demonstrates the applicability of our
method for general human actions. Fig.~\ref{fig:supp_interpolate} shows
interpolation along appearance and pose axes.

\subsection{Video Generation}
\label{imgc}
Requiring no pose annotations but only multiple images depicting the same object, our method can
be directly applied to video data without additional annotations. Thus, we are
also able to perform unsupervised video-to-video translation. We provide examples for
the norb dataset (\texttt{norb.avi}), the bbc dataset (\texttt{bbc.avi}) as well as the ntu dataset
(\texttt{ntu.avi}).

\newcommand{\cross}{\times}
\section{Implementation Details}
\label{impl}
In this section we provide additional details on the implementation of our method.
\subsection{Network Parameters}
We use the following notation to describe the network architectures:
\begin{itemize}
  \item \texttt{conv(n)}: a convolutional layer with $3\cross 3$-kernel and
    $n$ filters.
  \item \texttt{down(n)}: a convolutional layer with $3\cross 3$-kernel, $n$
    filters and stride $2$.
  \item \texttt{up(n)}: a convolutional layer with $3\cross 3$-kernel, $4n$
    filters, followed by a reshuffling of pixels to upsample the feature map
    by a factor of $2$. Also known as subpixel convolution \cite{subpixel}.
  \item \texttt{act}: a ReLU activation.
  \item \texttt{res}: a residual block \cite{resnet}: The input feature map plus the
    ReLU activated input feature map followed by a convolutional layer with $3\cross
    3$-kernel with as many filters as the input feature map has channels.
\end{itemize}
Depending on the dataset, the generated images have resolution $64\cross
64$ (sprites dataset), $96\cross 96$ (norb dataset) or $128\cross 128$
(remaining datasets). Both encoders $E_\pose$ and $E_\alpha$ have the same architecture:
\begin{itemize}
  \item Encoders at resolution $64\cross 64$: \texttt{conv(16), res, down(32), res, down(64), res,
    down(128), res, down(256), res, res, res, res, res, act, conv(16)}.
  \item Encoders at resolution $96\cross 96$: \texttt{
      conv(32), res,
      down(64), res,
      down(128), res,
      down(128), res,
      down(256), res,
      down(256), res,
      res, res, res, res, act, conv(16)}.
  \item Encoders at resolution $128\cross 128$: \texttt{
      conv(32), res,
      down(64), res,
      down(128), res,
      down(128), res,
      down(256), res,
      down(256), res,
      down(256), res,
      res, res, res, res, act, conv(16)}.
\end{itemize}
The decoder $D$ receives both $\pose$ and $\app$. Each of them is
processed seperately by four \texttt{res} blocks and the result is
concatenated. Depending on the resolution, the remaining decoder
architecture is described by:
\begin{itemize}
  \item Decoder at resolution $64\cross 64$: \texttt{
      conv(256),
      res, up(128),
      res, up(64),
      res, up(32),
      res, up(16),
      res, conv(3)
    }.
  \item Decoder at resolution $96\cross 96$: \texttt{
      conv(256),
      res, up(256),
      res, up(128),
      res, up(128),
      res, up(64),
      res, up(32),
      res, conv(3)
    }.
  \item Decoder at resolution $128\cross 128$: \texttt{
      conv(256),
      res, up(256),
      res, up(256),
      res, up(128),
      res, up(128),
      res, up(64),
      res, up(32),
      res, conv(3)
    }.
\end{itemize}
The classifiers $T$ and $T'$ have the same architecture. Both receive
$\pose$ and $\app$, and process them seperately with a \texttt{conv(512)} layer,
followed by four \texttt{res} blocks. The result is activated and processed
through a final \texttt{conv(512)} layer before the output is computed as
the inner product of the two resulting feature maps.

\subsection{Model Parameters}
For all experiments, we use $b_\gamma = l_\gamma = 10^{-2}$ and $\mu
= 10^{-1}$. $p(\pose \vert x_2)$ and $r(\pose)$ are both modeled as Gaussian
distributions of unit variance. The latter has a mean of zero and $E_\pose$
estimates the mean of the first. We use the reparameterization trick
\cite{VAE} to obtain low variance estimates of the gradient. Depending on
the dataset, we implement the negative log-likelihood $\Lrec$ with a $l_2$
loss (sprites), a perceptual loss \cite{perceptual} (norb) or a perceptual
loss together with a discriminator loss as in \cite{perceptualbrox},
weighted by $10^{-3}$ (remaining datasets).

\subsection{Optimization Parameters}
We train our model over batches of size $16$ for $100000$ steps. We
use the Adam optimizer \cite{adam} with an initial learning rate of
$2\cdot
10^{-4}$ linearly decayed to zero. We set $\beta_1=0.5$ and $\beta_2=0.9$.
$I_T$ is calculated with an exponential moving average with decay parameter 
$0.99$.

\newpage

\newcommand{\vufig}[1]{\raisebox{-.4\height}{\includegraphics[width=0.2\linewidth]{assets/suppmaterial/vucompare/#1}}}
\newcommand*{\myalign}[2]{\multicolumn{1}{#1}{#2}}
\newcommand*{\myhead}[1]{\makebox[32pt][c]{#1}}
\newcommand*{\myheader}{\myalign{c}{\myhead{}\myhead{ours}\myhead{vunet}}}

\begin{table}[tbp]
\begin{tabular}{lllll}
  &  \myheader & \myheader & \myheader & \myheader \\
  \toprule
  a) &  \vufig{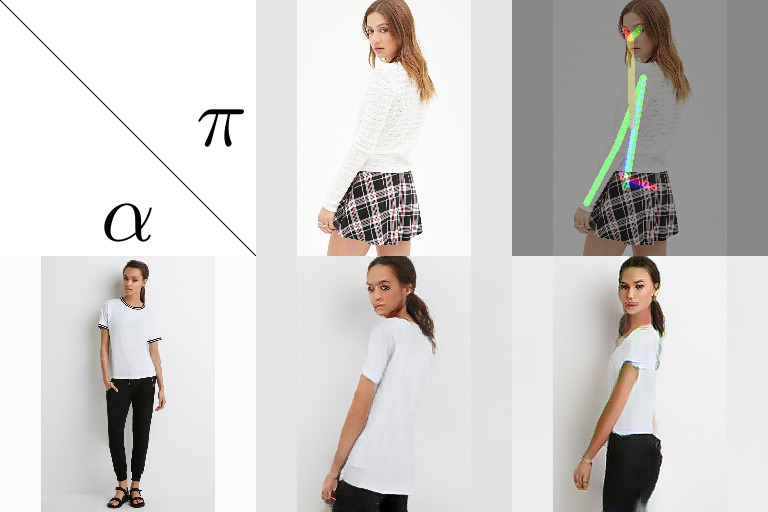} & \vufig{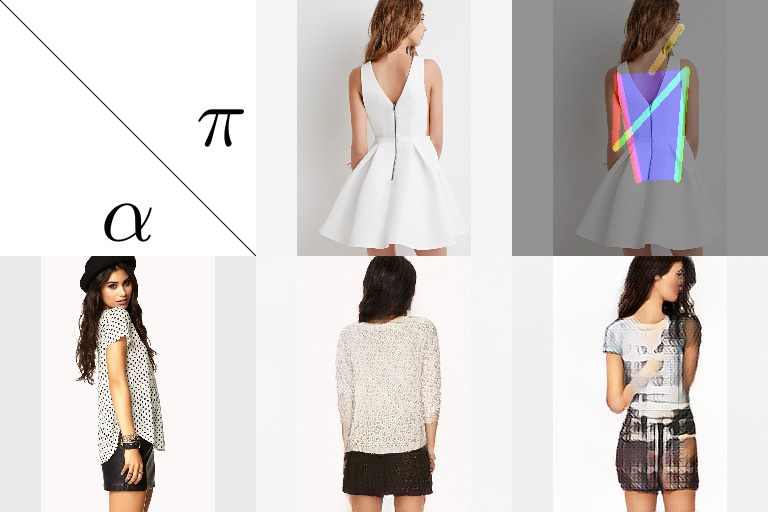} & \vufig{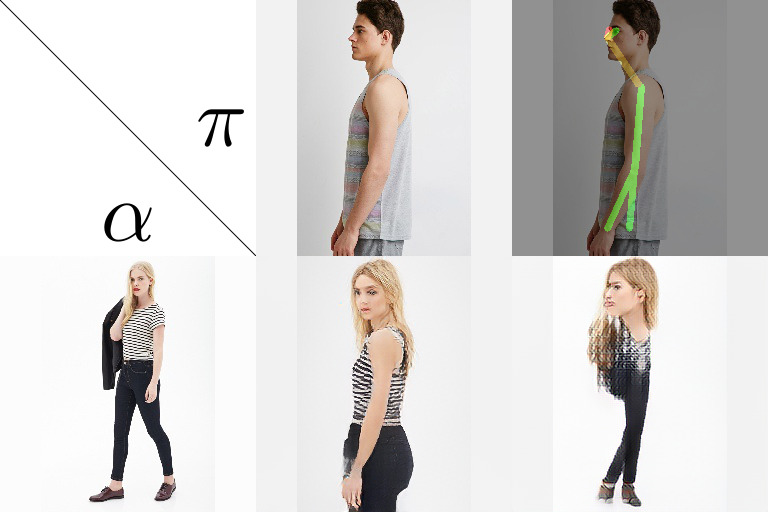} & \vufig{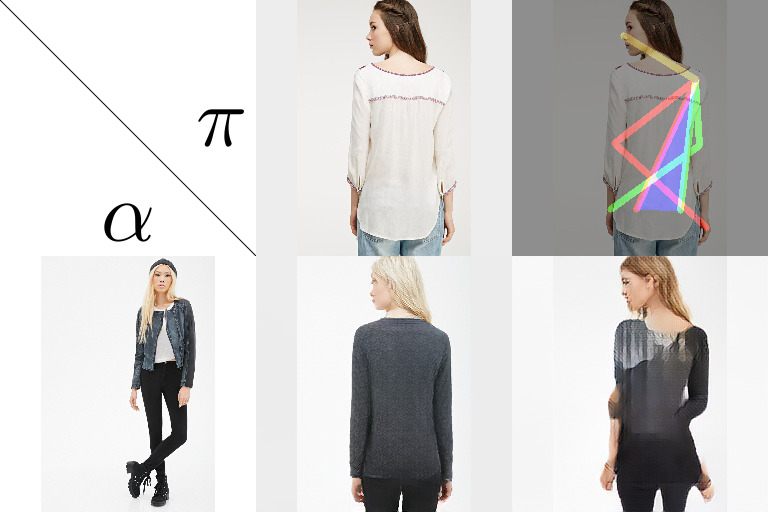} \\
  \midrule
  b) &  \vufig{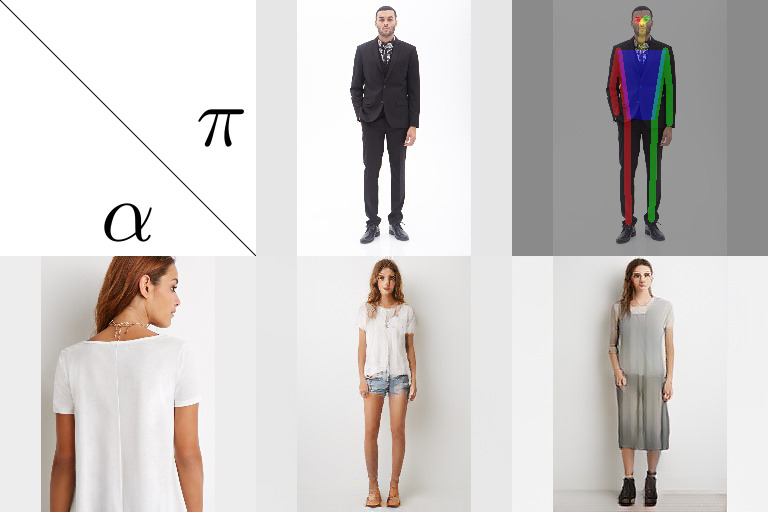} & \vufig{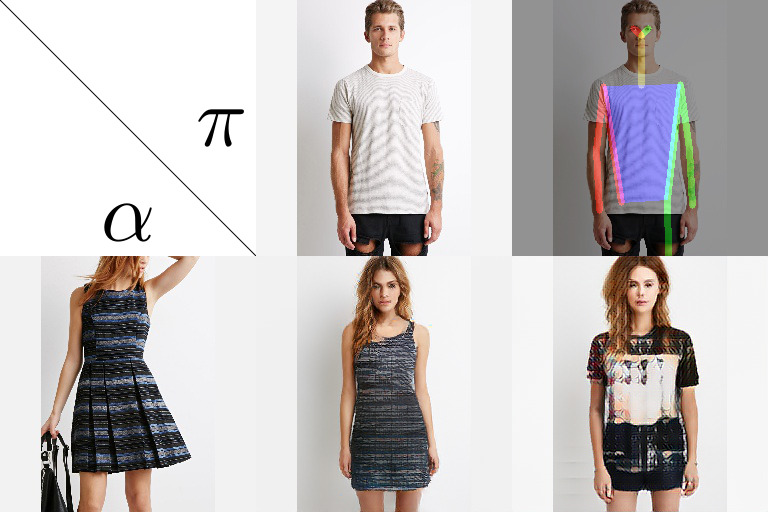} & \vufig{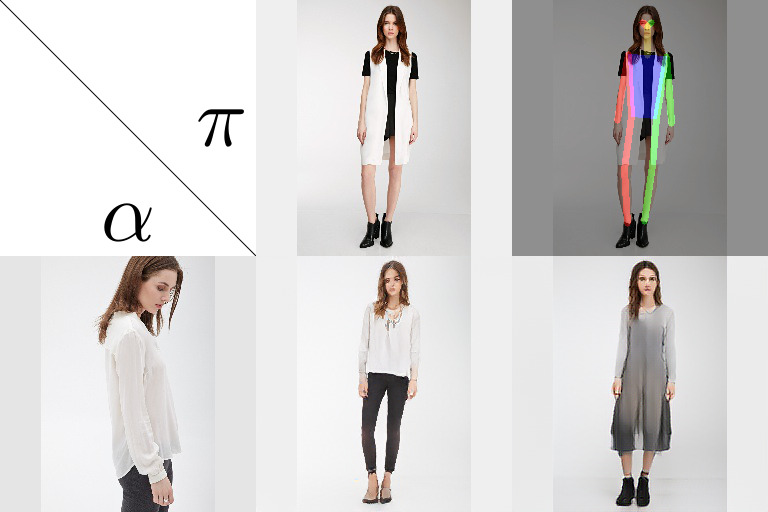} & \vufig{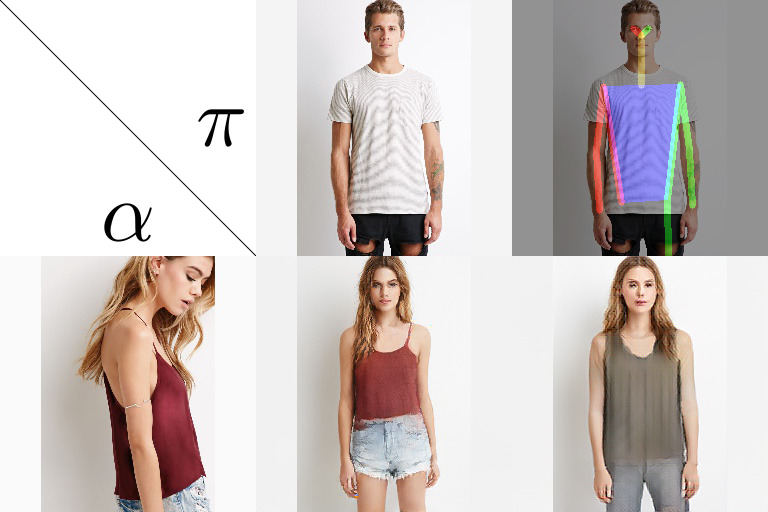} \\
  \midrule
  c) &  \vufig{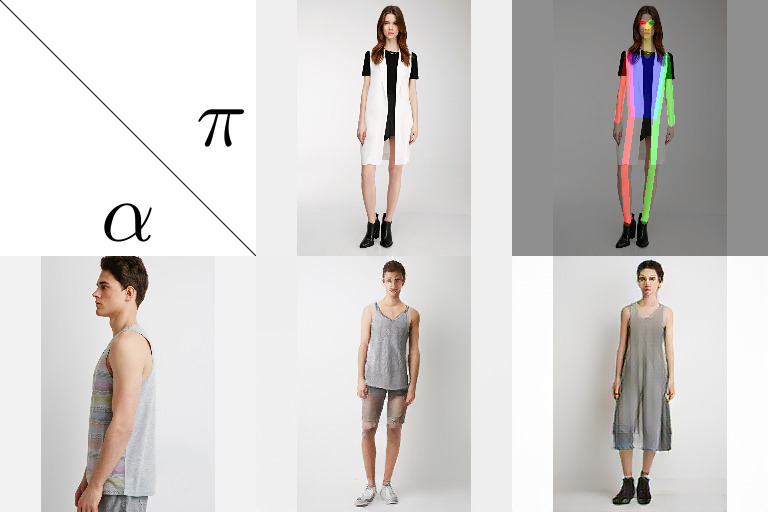} & \vufig{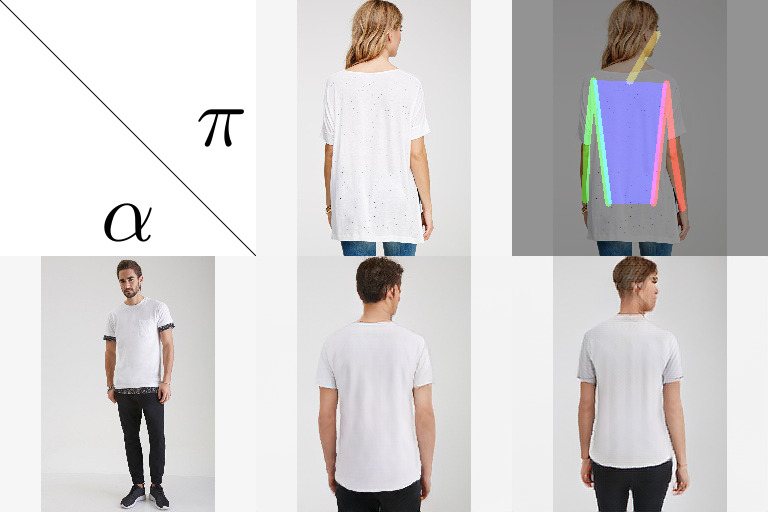} & \vufig{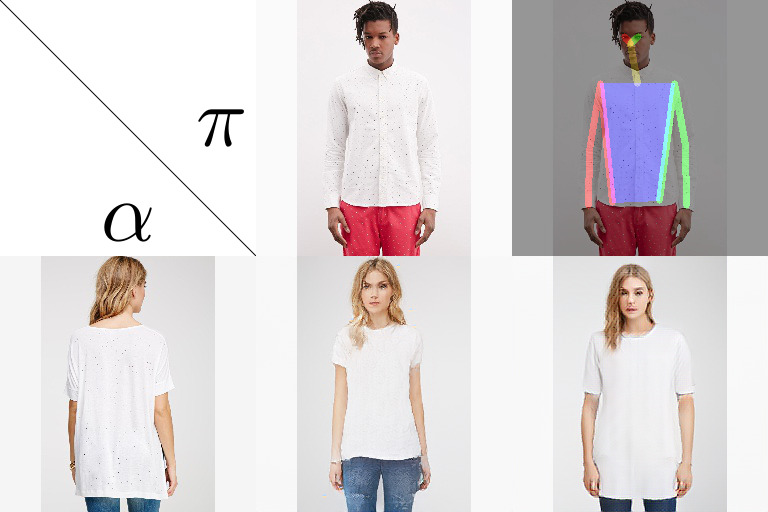} & \vufig{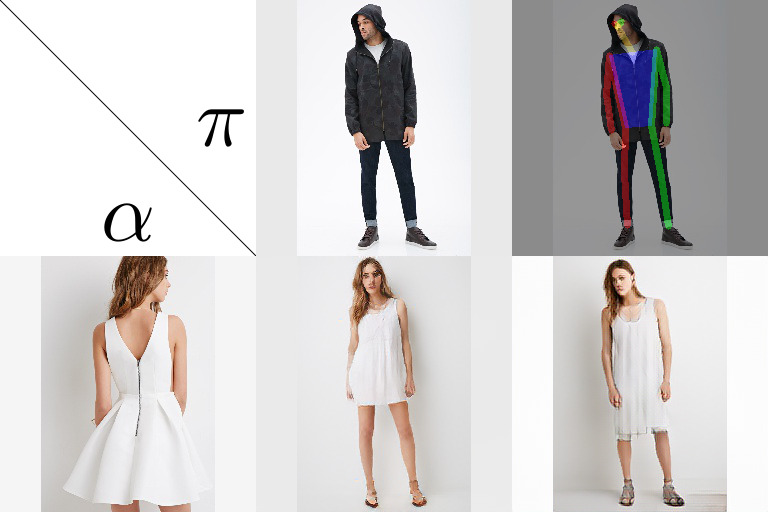} \\
  \bottomrule
\end{tabular}
  \caption{
    Comparison to vunet~\cite{VUNet}.
    Each matrix shows in the first row the pose target and the additional pose estimate used by vunet,
    and in the second row the appearance target followed by the synthesis of our method and vunet.
    a) vunet relies on existing pose estimators making it sensitive to
    estimation errors; our method always uses the direct target image for
    the pose.
    b) vunet also relies on pose estimates to obtain localized appearance
    representations, which can lead to complete failure at capturing the
    appearance.
    c) instead of learning a pose representation, vunet assumes that
    keypoints are good pose representations, but subtle information, e.g.
    shoulder width, still reveals information about appearances, e.g.
    gender. Men synthesized in poses estimated from women obtain a feminine
    look and vice versa. In contrast, our method is designed to learn
    completely disentangled representations and, in particular, learns
    gender-neutral pose representations.
  }
  \label{vucompare}
\end{table}

\newpage

\newpage

\newpage

\newpage

\newpage

\FloatBarrier

\end{document}